# Integrating Chain-of-Thought and Retrieval Augmented Generation Enhances Rare Disease Diagnosis from Clinical Notes

Zhanliang Wang[1,2#], Da Wu[1#], Quan Nguyen[1,3], Kai Wang[1,4*]

[1] Raymond G. Perelman Center for Cellular and Molecular Therapeutics, Children's Hospital of Philadelphia, Philadelphia, PA 19104, USA

[2] Applied Mathematics and Computational Science Graduate Program, University of Pennsylvania, Philadelphia, PA, 19104, USA

[3] Bioengineering Graduate Program, University of Pennsylvania, Philadelphia, PA 19104, USA

[4] Department of Pathology and Laboratory Medicine, Perelman School of Medicine, University of Pennsylvania, Philadelphia, PA 19104, USA

#: equal contribution

*: correspondence should be addressed to wangk@chop.edu.

**ABSTRACT**

**Background:** Several studies show that large language models (LLMs) struggle with phenotype-driven gene prioritization for rare diseases. These studies typically use Human Phenotype Ontology (HPO) terms to prompt foundation models like GPT and LLaMA to predict candidate genes. However, in real-world settings, foundation models are not optimized for domain-specific tasks like clinical diagnosis, yet inputs are unstructured clinical notes rather than standardized terms. How LLMs can be instructed to predict candidate genes or disease diagnosis from unstructured clinical notes remains a major challenge.

**Methods:** We introduce RAG-driven CoT and CoT-driven RAG, two methods that combine Chain-of-Thought (CoT) and Retrieval Augmented Generation (RAG) to analyze clinical notes. A five-question CoT protocol mimics expert reasoning, while RAG retrieves data from sources like HPO and OMIM (Online Mendelian Inheritance in Man). We evaluated these approaches on rare disease datasets, including 5,980 Phenopacket-derived notes, 255 literature-based narratives, and two cohorts of 1088 noisy clinical notes from a children's hospital**.**

**Results:** We found that recent foundations models, including Llama 3.3-70B-Instruct and DeepSeek-R1-Distill-Llama-70B, outperformed earlier versions such as Llama 2 and GPT-3.5. We also showed that RAG-driven CoT and CoT-driven RAG both outperform foundation models in candidate gene prioritization from clinical notes; in particular, both methods with DeepSeek backbone resulted in a top-10 gene accuracy of over 40% on Phenopacket-derived clinical notes, although performance was lower on noisier hospital clinical notes. RAG-driven CoT works better for high-quality notes, where early retrieval can anchor the subsequent reasoning steps in domain-specific evidence, while CoT-driven RAG has advantage when processing lengthy and noisy notes.

**Conclusions:** Integrating CoT and RAG enhances LLMs' understanding of clinical notes in the context of rare disease diagnosis,and can facilitate various downstream medical tasks.

## INTRODUCTION

Phenotype-driven gene prioritization and disease diagnosis is a systematic approach for identifying candidate causal genes or diseases based on clinical phenotypes observed in patients[1-4]. By matching observed patient phenotypes to known genotype-phenotype associations, this approach accelerates the discovery of pathogenic variants from whole genome/exome sequencing data, thereby facilitating timely and accurate clinical diagnosis of rare genetic disorders[5]. Effective phenotype-driven prioritization relies critically on structured annotations from standardized gene/phenotype databases, primarily including the Human Phenotype Ontology (HPO)[6], Online Mendelian Inheritance in Man (OMIM)[7], and Orphanet[8].Among these, HPO has become one of the most widely used resources due to its comprehensive vocabulary of phenotypic abnormalities, structured hierarchically as directed acyclic graph (DAG), enabling precise computational analyses, standardized data sharing, and interoperability across clinical and research settings[3,9].

Numerous computational tools leveraging structured HPO annotations have emerged in recent years, broadly categorized into two distinct methodological classes based on their prioritization logic and analytical framework[10].The first category, known as disease-first prioritization methods, initially ranks candidate disorders based on phenotype similarity before identifying the associated causative genes. Representative examples include: Phenomizer [11], which computes phenotype similarity scores; LIRICAL, which applies likelihood-ratio frameworks leveraging genotype-phenotype databases[12]; PhenIX[13] and Phen-Gen [14], which integrates phenotypic information with genomic data; Phrank, which employs hierarchical phenotype-set similarity measures [15]; and Phen2Disease, which uses bidirectional phenotype matching to prioritize diseases then genes[10]. The second category, termed direct gene prioritization methods, associates patient phenotypes directly to candidate genes without explicit disease ranking. Prominent examples include Phenolyzer, which integrates functional interaction networks [15]; Phen2Gene [16] , which is a probabilistic model that incorporates HPO annotations and gene prioritization; Exomiser[17], VarElect[18] and AMELIE[19], which combines phenotype similarity and genomic variant filtering; as well as other advanced graphical, machine learning or deep learning methods, such as PhenoApt[20], PhenoRank [21] , DeepPVP[22] , DeepSVP[23], GADO [24], and Xrare[25]. Phenotype annotations employed in these computational tools can be provided manually by clinicians or extracted automatically from clinical narratives through natural language processing (NLP) algorithms such as Doc2HPO[26] PhenoGPT and PhenoBCBert[27].

Despite significant advancements, current computational methods face several fundamental limitations. One significant limitation is that most existing methods require structured phenotype annotations, typically provided as standardized HPO terms. While NLP-based tools have automated phenotype extraction from clinical narratives, these methods inherently introduce an additional, cumbersome multi-step workflow: phenotype terms must first be extracted using NLP models, manually validated, and then separately provided as inputs to gene prioritization tools. Such a segmented approach increases complexity and reduces efficiency, limiting practical application into clinical workflows. Furthermore, most existing methods operate in a closed-system manner, relying solely on static curated genotype-phenotype databases while ignoring the broader clinical reasoning process. In contrast, human-clinicians frequently incorporate external medical references. Recent Retrieval-augmented methods, such as RAG-HPO, attempt to integrate external knowledge sources but may be limited by interpretability and retrieval granularity[27]. Additionally, clinical trust in these systems is hindered by their lack of transparent reasoning processes, as most methods function as black-boxes without explainable decision-making pathways[28]. Finally, current methods typically prioritize phenotype information alone, neglecting other clinically relevant patient-specific factors such a demographic information, laboratory test results, and family medical history, which could significantly enhance diagnostic precision and clinical relevance.

The swift advancements in Large Language Models (LLMs) significantly impact the biomedical research and clinical practice. By leveraging prior medical knowledge from pretrained LLMs[29], LLMs have the potential to serve as a powerful toolset for clinical decision support[30,31], patient education[32,33], medical questions answering, electronic health record (EHR) management, and personalized medicine[34,35]. While BERT-based models have been widely used in structured NLP tasks such as named entity recognition (NER) and relation extraction, they remain task-specific and require extensive fine-tuning. In contrast, GPT-based LLMs, such as GPT-series and Llama, demonstrate advanced reasoning and generative capabilities, allowing them to synthesize and contextualize complex biomedical data[36] . These strengths make them particularly valuable for disease diagnosis and gene prioritization, where they can analyze patient phenotypes, identify relevant genetic variants, suggest differential diagnoses, and generate evidence-based recommendations[35,37].

Several studies have already assessed the use of LLMs to perform disease diagnosis or gene prioritization tasks from phenotype data[38,39]. Models such as OpenAI GPT 3.5[40], GPT-4.0[41] and Llama 2 family[42] have struggled with these tasks. For instance, Kim et al.[38] found that even the best-performed LLM, GPT-4, achieved only average 17.0% accuracy in ranking diagnosed genes within the top 50 predictions, still falling short compared to traditional tools. This suggests that base LLMs may not have sufficiently learned high-quality, curated medical knowledge bases, likely due to their opaque nature and general-purpose training objectives. Besides that, standalone LLMs often hallucinate medical facts and lack access to real-time clinical knowledge, limiting their clinical reliability[35]. To address this, Retrieval-augmented generation (RAG)[43] allows LLMs to retrieve up-to-date biomedical knowledge from curated databases and genetic repositories, improving the accuracy of diagnosis[43] . Recent advances have introduced dynamic retrieval methods such as Self-RAG[44] and adaptive reranking strategies[45], as well as multimodal RAG frameworks that integrate text with medical images[46,47]. However, these often require complex infrastructure or multimodal datasets not readily available for rare disease diagnosis, where clinical documentation remains predominantly text-based. Additionally, Chain-of-Thought (CoT)[48] reasoning enhances interpretability by enabling models to explicitly outline their diagnostic logic, making AI-assisted predictions more transparent and clinically useful[49]. Zelin et al.[50] explored these approaches in rare disease diagnosis, and found that RAG alone led to a slight improvement, increasing accuracy from 37% with the base ChatGPT-3.5 to 40%. However, adding zero-shot CoT did not provide any further benefits, as performance remained at 40%. Additionally, applying CoT alone to the base model reduced accuracy to 23%, suggesting that without retrieval, CoT may introduce reasoning biases.

While these existing approaches require minimal training, they have notable limitations. The effectiveness of the RAG depends heavily on the quality and comprehensiveness of the underlying database, which can limit its ability to retrieve accurate and relevant biomedical knowledge. Additionally, several studies show that neither RAG nor prompt engineering techniques consistently lead to significant performance improvements. Sometimes simply combining multiple methods such as RAG and CoT does not always enhance accuracy, and in some cases can even reduce prediction performance and hallucinate on medical facts[38,39].

In this study, we aim to investigate whether LLMs can perform fully automated gene prioritization and disease diagnosis directly from clinical notes in a single step, eliminating the need for an intermediate step that converts clinical data into standard terminologies such as HPO, PhecodeX[51], SNOMED-CT[52] or Phenopacket[53]`. Achieving this level of automation is essential for making LLMs more practical in real-world clinical settings by reducing reliance on manual data preprocessing and ensuring seamless integration into diagnostic workflows.

Beyond this core objective, we explore several key questions to better understand the behavior of LLMs and improve prediction accuracy in this domain. First, we examine whether additional clinical information, such as patient demographics, age, gender, lab results and family history can provide additional clues to LLMs to infer candidate genes or candidate diseases. Second, while disease prioritization has been explored in previous studies[39] , our approach uses a more advanced and reasonable evaluation method to improve accuracy and robustness. Instead of relying solely on exact match to assess performance, we compare predictions based on semantic similarity match, ensuring that the synonyms of the same disease are correctly accounted for. This method reduces risk of both false positives and false negatives, making it more effective in handling the complexity of disease labels. Finally, we assess the effectiveness of modern enhancement techniques designed for LLMs, such as CoT and RAG, in improving the domain-specific inference for disease and gene prioritization. Unlike recent complex multimodal RAG approaches[46], our methods focus on text-based clinical notes, the predominant data format in rare disease diagnosis, optimizing the integration order between CoT and RAG. To enhance performance while ensuring that the diagnostic process remains interpretable, aligning more closely with human clinicians' reasoning, our goal is to strategically integrate RAG and CoT to maximize their complementary strengths. To achieve this objective of interpretability, we present two hybrid strategies: RAG-driven CoT and CoT-driven RAG. These methods are designed to refine the synergy between retrieval-based augmentation and logical reasoning while requiring no additional training. We demonstrate that these methods can improve the performance of base LLMs (Llama 3.3-70B-Instruct[54] and DeepSeek-R1-Distill-Llama-70B[55]) in gene prioritization and the rare disease diagnosis from noisy and lengthy free-text clinical notes. We also include OpenAI GPT-4o[56] (Feb. 2025 version) in our case studies, further highlighting the effectiveness of CoT prompting regardless of what base model is used.

## RESULTS

### *Summary of Datasets and LLMs*

In this study, we assess the performance of various LLMs on rare genetic disorders diagnosis and gene prioritization, using four distinct test datasets (**Table 1**): 5,980 Phenopacket-derived clinical notes, 255 clinical narratives extracted from PubMed articles, 220 in-house clinical notes from EHRs, and an Arcus clinical-notes dataset from the Children's Hospital of Philadelphia (CHOP). The "Phenopacket-derived Clinical Notes" (**Table 1**) dataset comprises notes derived from structured Phenopacket-store data[57]. The original Phenopacket-store data included information such as age, sex and HPO IDs in structured JSON formats, which we used to prompt GPT-4 to generate extensive clinical notes (The prompt used for generating synthetic clinical notes is presented in **Table S1**. Note that to conserve computational resources and ensure a more balanced dataset, we limit our selection to a maximum of two patients per disease/gene. The "PubMed Free Text" dataset includes 255 free text paragraphs detailing patient demographics and HPO terms sourced from PubMed publications. We also gathered the "In-house Clinical Notes" from the Children's Hospital of Philadelphia (CHOP) to test the methods on real clinical notes that are much noisier and lengthier than notes from scientific publications. Similarly, the Arcus dataset contains noisy free-text clinical notes and serves as a semi-independent cohort (868 cases for disease diagnosis; 702 cases for gene prioritization), enabling us to evaluate whether the same prompt or retrieval strategies work on a different cohort. Arcus is an information platform providing access to clinical and research data in a customized computing space hosted at Google, and may contain patients with positive genetic diagnostic performed by outside diagnostic labs. Altogether, these notes were compiled from patients affected with a wide range of rare genetic diseases with confirmed diagnoses from genetic testing.

Our main analysis focused on two foundation models: Llama 3.3-70B-Instruct and DeepSeek-R1-Distill-Llama-70B, and we examined different strategies to utilize foundation models by integrating CoT and RAG into the inference process. For each dataset, we employed four different methodologies to predict rare genetic diseases: (1) direct inference using the base prompt, (2) direct inference with the Chain of Thought (CoT) prompt, (3) direct inference with the base prompt enhanced by Retrieval-Augmented Generation (RAG), (4) RAG-driven CoT, and (5) CoT-driven RAG. Both RAG-driven CoT and CoT-driven RAG integrate RAG with CoT, but they differ in the order that each component is applied. In the former case, the RAG system is activated to perform searches before the model engages in structured thinking (CoT). Conversely, in the latter case, the model first undertakes structured thinking guided by CoT and then uses the structured outputs (such as HPO terms) to query the RAG database. The graphical illustration of RAG-based methods, including (3), (4) and (5), are presented in **Figure 1**. Finally, while we did not conduct a full-scale evaluation of OpenAI GPT-4o (Feb. 2025 version), we include one case study showcasing the power of CoT prompting.

### ***Accuracy of LLMs on Rare Disease Diagnosis and Gene Prioritization***

In the subsections below, we summarize the accuracy of various flavors of LLMs on the task of performing rare disease diagnosis and candidate gene prioritization from clinical notes directly. The primary results for Top-10 accuracy on the three testing datasets are displayed in **Figure 2**, while **Table**

**S2** provides a comprehensive breakdown of both Top-1 and Top-10 accuracies for all models and inference strategies.

Base Model Performance

Both the latest open-source LLMs (Llama3.3-70B-Instruct and DeepSeek-distill-Llama3) demonstrate reasonable base line performance across various datasets and tasks. In PubMed Free-text gene prioritization task, for instance, Llama3.3-70B-Instruct achieves a Top-10 accuracy of 27.84% and a Top-1 accuracy of 16.47%, while DeepSeek-distill-Llama3's baseline stands at Top-10 of 21.57% and Top-1 of 7.06%. On the Phenopacket-derived clinical notes with higher quality, baseline Top-10 accuracy for Llama3.3-70B-Instruct ranges from 32.68% to 35.03% for disease or gene prioritization tasks, with the corresponding Top-1 accuracy hovering around 20%. DeepSeek-distill-Llama3, by contrast, has lower accuracy for gene prioritization (11.72% Top-10, 5.68% Top-1) but performs competitively in disease diagnosis (31.65% Top-10, 20.37% Top-1). Both methods on the In-house clinical notes, which are notably noisier than the other two datasets, exhibit somewhat variable baseline performance: around 30% in Top-10 for disease diagnosis and 20% for gene prioritization, reflecting the unstructured complexity of real-world clinical documents. On the Arcus dataset, baseline performance was lower: Llama 3.3-70B-Instruct achieved 22.00% Top-10 (8.18% Top-1) for disease diagnosis and 9.40% Top-10 (4.98% Top-1) for gene prioritization, while DeepSeek-R1-Distill-Llama-70B achieved 21.19% Top-10 (11.18% Top-1) for disease diagnosis and 5.98% Top-10 (4.13% Top-1) for gene prioritization.

Chain-of-Though (CoT) Prompting Performance

CoT prompting, where the model explicitly manifests its intermediate reasoning steps, consistently boosts both Top-1 and Top-10 accuracy across distinct tasks, though gains differ by datasets. In PubMed Free-text gene prioritization task, for example. Llama-3.3-70B-Instruct's Top-10 rises from 27.84% to 30.58%, with Top-1 hovering near 16-17%. However, the Phenopacket-derived tasks show larger CoT-driven improvements, for instance, DeepSeek-distill-Llama3's gene prioritization accuracy surges from 11.72% to 41.18% in Top-10, with Top-1 from 5.68% to beyond 20%. Disease diagnosis similarly benefits from CoT, particularly when phenotypic descriptions are systematic and allow the model to methodically connect clinical features with candidate diseases. Even the more challenging in-house data often sees a 4-5% lift in Top-10 when CoT is applied, suggesting that structured reasoning can mitigate influences of noises in real-world clinical notes.

Retrieval-augmented Generation (RAG) Performance

By giving the model access to external biomedical resources, RAG addresses knowledge gaps not covered by either pretraining or model's thinking process. This is especially evident in high-quality dataset with richer phenotypic information: DeepSeek-distill-Llama3's disease diagnosis on Phenopacket-derived clinical notes jumps from 31.65% (Top-10) to nearly 37.87% under RAG, and Llama3.3-70B-Instruct similarly improves as well. We note that RAG can yield more variable outcomes in PubMed Free-text or in-house contexts, where retrieved information sometimes introduces noise or irrelevant details, thus degrading performance. Nonetheless, improvements in Top-1 accuracy in some conditions confirm that plugging key knowledge gaps sharpens the model's capacity to identify the correct disease or gene first.

RAG-driven CoT Performance

In this hybrid approach, retrieval precedes the Chain-of-Thought, ensuring that any newly fetched references guide the reasoning from the outset. For Phenopacket-derived tasks, especially gene prioritization task, such early context can yield the highest Top-10 accuracies: DeepSeek-distill-Llama3 ascends from 11.72% to 42.13% for Top-10 accuracy, and similarly from 5.68% to 23.78% for Top-1 accuracy. Llama3.3-70B-Instruct shows a similar trend in Phenopacket-derived disease diagnosis, increasing from 32.68% to 37.86% in Top-10 accuracy and 20.74% to 24.97% in Top-1 accuracy. In contract, for PubMed Free-text, RAG-driven CoT affords only modest lifts, (e.g., only 27.84% to 28.24% in Llama3.3-70B-Instruct's Top-10), and in in-house data it can even underperform CoT alone if retrieved content is not tightly focused (e.g., Llama3.3-70B-Instruct's disease diagnosis task falls short of CoT's 35.00% Top-10 at just 27.73%).

CoT-driven RAG Performance

CoT-driven RAG instead has the model clarify its reasoning before retrieval, thus exposing the precise information gaps that external sources should fill. In Phenopacket-derived settings, CoT-driven RAG remains competitive. For instance, DeepSeek-distill-Llama3's gene prioritization task reaches Top-10 accuracy of 41.54%, only marginally below the best value of 42.13% (RAG-driven CoT), while Top-1 stands at 23.20%.

The more challenging in-house data generally favors this reasoning-first approach, as extensive irrelevant text can mess up the retrieval process, if performed prematurely. DeepSeek-distill-Llama3's disease diagnosis accuracy climbs from 29.55% to 35.00%, substantially higher than the 28.18% achieved by RAG-driven CoT, and a similar trend holds for gene prioritization (from 8.18% to 25.91%). Llama3.3-distill-Llama3's disease diagnosis accuracy likewise benefits, moving from 30.45% to 34.09% under CoT-driven RAG, compared to 27.73% with RAG-driven CoT. We observed the same trend on the Arcus cohort, which also consists of noisy clinical notes: CoT-driven RAG achieved the strongest overall performance on Arcus clinical notes for both disease diagnosis and gene prioritization (Llama 3.3-70B-Instruct: 30.30% Top-10 for disease and 19.09% for gene; DeepSeek-R1-Distill-Llama-70B: 31.57% Top-10 for disease and 18.67% for gene), supporting the robustness of the reasoning-first retrieval strategy under documentation noise.

Comparative Observations and Clinical Relevance

Across the four datasets, PubMed Free-text, Phenopacket-derived clinical notes, In-house clinical notes, and Arcus clinical notes, and for both rare disease diagnosis and gene prioritization tasks, our evaluations reveal that Chain-of-Thought (CoT) prompting and Retrieval-augmented Generation (RAG) each offer substantial gains over baseline performance. Although CoT alone often delivers notable improvements by clarifying how phenotypic details map to diseases or genes, RAG provides complementary advantages where targeted, up-to-date knowledge is lacking. Their synergy emerges most vividly in hybrid approaches, RAG-driven CoT and CoT-driven RAG, whose relative efficacy depends on the complexity and structure of the source data. In relatively high-quality clinical notes such as Phenopacket-derived clinical notes, RAG-driven CoT can maximize Top-10 accuracy by grounding the model's reasoning in domain-specific references from the outset. Conversely, noisy or lengthy datasets such as in-house clinical notes and Arcus Genetics clinical notes often benefit more from CoT-driven RAG, where an initial Chain-of-Thought narrows the search space and prevents extraneous information from damaging final predictions. Taken together, these results underscore the versatility of combining explicit reasoning with selective knowledge retrieval to enhance model accuracy in clinical scenarios such as disease diagnosis and gene prioritization. Statistical significance testing on Top-1 accuracy using two-proportion z-tests (binomial approximation; Table S3 further supports these observations. For Llama 3.3, CoT-driven RAG significantly outperformed RAG on PubMed Free Text (gene) ($p = 0.035$) and on in-house clinical notes (gene) ($p = 0.020$), consistent with the benefit of coupling reasoning with retrieval in settings where retrieval alone may introduce distracting context. In contrast, most other pairwise comparisons did not reach statistical significance ($p \geq 0.05$), suggesting that while the direction of improvement is often consistent, effect sizes can be modest and limited by dataset-specific sample sizes.

***Semantic Embedding Space Analysis of Different RAG-based Approaches***

To understand the semantic relationships within our knowledge base, we projected high-dimensional embedding (768-dimension) of query vectors and all 64,068 chunked documents from the HPO and OMIM databases into a two-dimensional space using PaCMAP[58]. Chunked documents refer to breaking down long text into smaller, manageable pieces ("chunks") before feeding them into LLMs for processing, and this procedure is applied on texts retrieved from both the HPO and OMIM databases. Query vectors, which encode input clinical notes or intermediate reasoning steps, provide latent representations of the model's input. As embeddings encode semantic meaning, closer distances in the 2D space reflect semantic similarity between documents. Our experimental setup derives query vectors from all three datasets (Phenopacket-derived clinical notes, PubMed Free-text, and in-house clinical notes).

As shown in **Figure 3**, the two-dimensional projection of our chunked documents reveals three distinct clusters labeled I, II, and III. **Cluster I**, which is composed of 99% HPO documents and 1% OMIM documents, forms a compact group. The documents in this cluster exhibit standardized formats with clear field identifiers and a systematic organization of phenotypic information by organ system; for instance, entries neatly catalog neurological and cardiovascular features in a list-based format. In contrast, the OMIM documents form two separate clusters. One of these clusters, **Cluster II**, which is composed of 95.5% OMIM and 4.5% HPO documents, is dominated by clinical-descriptive texts that offer detailed case narratives, including patient age, geographic context, and syndrome classifications. The other OMIM cluster, **Cluster III,** which is composed of 100% OMIM documents, focuses on molecular-mechanistic details—presenting information on specific proteins such as glutathione peroxidase and neuromedin U, along with experimental findings that elucidate biochemical processes. Together, these clusters highlight the clear semantic distinctions between the highly structured, phenotype-taxonomic HPO entries and the more diverse, narrative-rich OMIM documents.

**Figure 3b** illustrates the placement of query vectors generated using the RAG-driven CoT approach. These embeddings are centrally located within **Cluster II**. This positioning suggests that the RAG-driven CoT method is designed to initially retrieve a broad spectrum of clinically rich documents. This approach enables the system to gather diverse clinical insights before constructing a refined Chain-of-Thought that ultimately synthesizes the final answer.

In contrast, **Figure 3c** presents the query vectors from the CoT-driven RAG approach. Here, the queries are located much closer to the **Cluster I** (HPO cluster). This positioning indicates that when a Chain-of-Thought is generated prior to retrieval, the process prefers the more concise, phenotype-oriented HPO documents. The prior reasoning effectively filters the retrieval, leading to a more focused set of documents that emphasize key clinical features.

The spatial arrangement underscores the influence of query quality on retrieval effectiveness in the RAG procedure. The heterogeneity of the query vectors stemming from Phenopacket-derived clinical notes, PubMed Free-text, and in-house data, demonstrates that even though these queries come from varied sources, their ultimate semantic alignment with the RAG database is governed by the inherent characteristics of HPO and OMIM documents. Consequently, retrieval effectiveness is determined by the degree to which the semantic content of the query corresponds to either the structured, phenotype-oriented nature of HPO or broader, more detailed style of OMIM.

To better understand how the chunk size of HPO and OMIM documents might affect retrieval outcomes, we also analyzed the chunked document length distribution in the knowledge base for retrieval as shown in **Figure 3**. The combined RAG knowledge composed by HPO and OMIM database contains a total of 64,068 chunked documents, with document lengths ranging from 5 to 511 tokens. The mean length is 282.9 tokens while the median length is 285.0 tokens, indicating a relatively balanced distribution of chunked document sizes. In addition, the HPO Database contributes 15,762 chunks, while the OMIM Database provides 48,136 chunks. **Figure 3d** shows a density curve with multiple peaks, suggesting that different sources contribute to varied document length with clusters around 150, 280, and 450 tokens. In **Figure 3e and 3f**, the length distributions for HPO and OMIM are displayed separately, revealing that HPO chunks tends to be shorter and more uniform (mean: 210.98 tokens; median 218.00 tokens), whereas OMIM spans a broader range of lengths (mean: 307.65 tokens; median: 333.50 tokens). These differences in chunk length can influence retrieval effectiveness: shorter documents may yield direct and highly focused matches, while longer documents offer more context but risk introducing extraneous details.

The comprehensive analysis of the semantic embedding space and document-length distributions shows that query vectors from different RAG strategies (RAG-driven CoT vs. CoT-driven RAG) interact with the heterogeneous HPO-OMIM knowledge base in markedly different ways. The order of retrieval and chain-of-thought construction exerts a strong influence on where queries land in the embedding space. Specifically, queries under RAG-driven CoT tend to gravitate toward the OMIM clusters, thereby retrieving more detailed case reports and molecular discussions, whereas CoT-driven RAG queries often align with the structured, phenotypic-oriented HPO content. Furthermore, the inherent differences in document lengths, characterized by shorter, standardized HPO chunks as opposed to longer, narrative-rich OMIM chunks, play a critical role in determining the granularity of retrieval outcomes. Shorter chunks tend to provide more precise matches while longer chunks tend to provide a broader context.

### *Case Studies on the Impacts of CoT and RAG-based Techniques*

In this section, we present two case studies – involving ZIC3 and ACTA1 as the causative genes. These two distinct case studies were conducted to illustrate how our system performs under varying technical complexities. In the ACAT1 case, we assessed the full spectrum of RAG-based pipelines including RAG, RAG-driven CoT, CoT-driven RAG, demonstrating their effectiveness for gene prioritization when substantial retrieval and structured reasoning are required. By contrast, the ZIC3 case was designed to highlight a lower-overhead workflow such as ChatGPT web interface, that relies on a simpler CoT approach, making it more accessible to users with minimal technical expertise.

Throughout both cases, Llama3.3-70B-Instruct and DeepSeek-R1-distill-Llama3 yielded broadly similar gene rankings, so we present primarily Llama3.3-70-Instruct's outputs are representative examples. Full response given by various models are provided in the Supplementary Materials.

#### *Case 1: RAG-based Methods Enhances ACTA1 Gene Prioritization*

We first assessed clinical notes on a female patient affected with a range of clinical phenotypes (**Table 3**), who carries a variant in ACTA1 as the diagnostic variant through whole exome sequencing. The ACTA1 gene encodes skeletal muscle alpha-actin. Mutations in the ACTA1 gene are linked to various congenital myopathies, including nemaline myopathy and intranuclear rod myopathy[59-61]. Disease severity varies widely, from neonatal respiratory failure to mild, adult-compatible forms.

With the base Llama 3.3-70B-Instruct, the true label is ranked 6th among the Top-10 predictions. The model prioritized MTM1, BIN1, TNNT1, TPM2 and NEB as the top 5 genes, which are genes that, while associated with neuromuscular disorders[62], either follow different inheritance patterns or lack a full phenotype match. This result is consistent with prior studies showing that LLMs struggle with phenotype-to-gene mapping due to the lack of structured medical reasoning[63].

By incorporating retrieval augmentation before structured reasoning, the RAG-driven CoT successfully prioritized ACTA1 as the top-ranked gene, with the final top ranked genes being ACTA1, NEB, TPM2, TPM3, TNNT1, KBTBD13, KLHL40, KLHL41, LMOD3, MYH7, respectively. This demonstrates the effectiveness of retrieval augmentation, which brought in high-quality external knowledge, leading to improved ranking. We note that the strategy introduced some genes with incomplete phenotype overlap with the disease such as TPM3, KLHL40, and LMOD3. As rare genetic diseases tend to have high phenotypic heterogeneity, it is expected that complete phenotype matching with what is documented in HPO/OMIM is unlikely for any given patient. In any case, these findings align with recent studies showing that RAG-enhanced LLMs can retrieve gene-related information effectively but may introduce external noise as well[64].

In contrast, CoT-driven RAG first structured the extracted phenotypic features before retrieval, aligning the reasoning process more closely with human clinical workflows. The structured feature extraction process ensured that only highly relevant genes were retrieved. The final Top-10 prioritized genes includes ACTA1, RYR1, TPM2, KBTBD13, BIN1, DNM2, MYPN, NEB, TNNT1, and CCDC174, which aligns more closely with expectations from human experts. This finding demonstrates that CoT-driven RAG effectively filters out irrelevant genes while maintaining high retrieval accuracy.

*Case 2: Chain-of-Thought Prompting Enhances ZIC3 Gene Prioritization*

We next assessed clinical notes on a male patient affected with multiple congenital cardiac anomalies and splenic abnormalities (**Table 3**), who carries a variant in ZIC3 as the diagnostic variant through whole exome sequencing. The ZIC3 gene encodes a putative zinc finger transcription factor[65]. ZIC3 mutations are associated with X-linked heterotaxy, a disorder characterized by disruptions in embryonic laterality and midline developmental field defect[66-68].

[69]When using the Base Llama3.3-70B-Instruct model, ZIC3 ranked only fourth among the top ten genes, with other cardiac-development genes (e.g., NKX2-5, GATA4, TBX5) placed higher. Similarly, the closed source OpenAI-GPT4o model (base configuration) listed ZIC3 in third place suggesting that although these foundational LLMs recognized its relevance, they did not consistently prioritize it as the top candidate.

By contrast, introducing Chain-of-Thought (CoT) prompting elevated ZIC3 to the first rank for both Llama3.3-70B-Instruct and OpenAI-GPT4o. In the CoT-driven outputs, the models explicitly extracted and classified key phenotypic features such as congenital heart anomalies under the "Cardiovascular system", polysplenia under the "Immune system". This structured approach filtered out less specific cardiac-development genes that lacked a complete phenotype match for the patient's laterality defects[70,71], leading each CoT-based model to pinpoint ZIC3 as the most likely causative gene.

## DISCUSSION

In this study, we demonstrated the potential of Large Language Models (LLMs) for automated gene prioritization and rare disease diagnosis directly from free-text clinical notes, eliminating the need for structured phenotype annotations such as HPO terms. We also introduced RAG-driven CoT and CoT-driven RAG, two methods that combine CoT and RAG to analyze clinical notes, and show that they have improved performance over foundation models. The prioritized gene names or disease names may be used in conjunction with other information, such as whole genome/exome sequencing data, to facilitate the identification of disease causal genes, or may be useful in selecting the appropriate diagnostic modalities/strategies, such as repeat expansion assays. We concluded that integrating CoT and RAG enhances LLMs' understanding of clinical notes, and can facilitate downstream medical tasks. Below we discuss several important technical issues related to model performance and propose future directions for additional improvements.

First, base LLMs demonstrated reasonable but suboptimal accuracy in phenotype-driven gene prioritization and disease diagnosis. For example, on the Phenopacket-derived data, Llama 3.3-70B-Instruct achieved Top-10 accuracies of 32.68% for disease diagnosis and 35.03% for gene prioritization, while DeepSeek-R1-distill-Llama3 scored 31.65% for disease diagnosis and 11.72% for gene prioritization. These performances surpassed older model such as Llama2 and OpenAI's GPT-3.5, indicating that the latest open-source LLMs have internalized a significant amount of medical knowledge. However, the lower accuracy for DeepSeek-R1-distill-Llama3 on gene prioritization suggests that DeekSeek may have limited knowledge in phenotype-genotype mapping due to inherent biases in training data. The development of modern LLMs such as DeepSeek usually involves extensive pre-training on vast text corpora followed by careful preference alignment with human feedback. While these large pretrained models are incredibly powerful, they are somewhat opaque, making it difficult to determine or quantify the specific knowledge they acquire during pre-training. For instance, the extent of their understanding of rare diseases and their reliability for clinical use remains uncertain. In fact, previous research[72] has shown that LLMs generally tend to predict commonly occurred disease names. These challenges, coupled with the "black box" nature of generic pretraining, have driven us to develop and test various techniques that can steer LLMs towards more reliable predictions in these domain-specific tasks.

Second, Chain-of-Thought (CoT) reasoning led to substantial accuracy improvements, though its effectiveness is variable across datasets or base LLMs. For instance, when applied to DeepSeek-R1-Distill-Llama3, CoT prompting increased gene prioritization accuracy from 11.72% to 41.18%, demonstrating that structured reasoning can greatly enhance phenotype-to-gene inference. In comparison, the disease diagnosis accuracy for Llama3.3-70B-Instruct improved from 30.45% to 35.00% on the in-house dataset through CoT prompting. This variability is likely due to the ongoing enhancements in the reasoning capabilities of LLMs[55,73-76], different tolerance of noises in the input documents, and the differences in pretraining data of base LLMs as previously mentioned. Our manual investigation of in-house clinical notes also indicates that notes with longer contexts and noisier inputs gain greater advantages from CoT prompting generally. In summary, prompted CoT reasoning helps refine predictions by guiding the model toward more structured decision-making processes, making it particularly useful for complex clinical inference tasks.

Third, Retrieval-augmented generation (RAG) strategies provided additional accuracy gains by integrating external biomedical knowledge base such as the HPO database and OMIM database. By retrieving structured and unstructured information from these databases, RAG-enhanced models were able to reduce hallucinations and improve gene-disease association mapping. However, we also found that RAG-driven CoT and CoT-driven RAG differed in effectiveness: (1) RAG-driven CoT, which retrieves external knowledge before applying structured reasoning, lead to moderate accuracy improvements across all datasets. (2) CoT-driven RAG, which first applies structured reasoning and then retrieves supporting knowledge, consistently outperformed all other methods. For example, in Phenopacket-derived clinical notes for gene prioritization, CoT-driven RAG improved Top-10 accuracy for DeepSeek-R1-distill-Llama3 from 11.72% to 41.54%, a substantial improvement. Similarly, on in-house dataset, CoT-driven RAG can bring the accuracy from 6.36% to 14.09%, when applied to the DeepSeek-R1-distill-Llama3 model. These results suggest that applying structured reasoning before retrieval helps refine model queries, leading to more relevant knowledge retrieval and improved diagnostic accuracy.

Fourth, we note that both RAG-driven CoT and CoT-driven RAG differ from the Retrieval Augmented Thoughts (RAT) approach described in existing literature[77], where RAT methodically revises the sequential thoughts generated by a zero-shot CoT prompt applied to an LLM. While RAT is not formally evaluated in this paper, our private tests indicate that its performance is comparable to that of RAG-driven CoT and CoT-driven RAG. However, RAT requires significantly more inference time with lengthy clinical notes, leading us to conclude that it is less of a practical option. Recent research has explored various RAG optimization strategies for biomedical applications. Adaptive retrieval methods such as Self-RAG[44] enable models to dynamically determine when and how much information to retrieve based on query complexity, while dynamic reranking approaches[45] refine retrieval results to improve relevance. Multimodal RAG frameworks extend beyond text to incorporate medical images; for example, MMed-RAG[46] integrates vision-language models with domain-aware retrieval for medical visual question answering, and RULE[47] introduces calibrated context selection for factuality control. Table S1 compares these approaches with our methods based on design characteristics. While these advances show promise, multimodal approaches require paired image-text datasets not readily available for rare disease diagnosis, where clinical documentation remains predominantly text-based. Our methods address this complementary use case by optimizing reasoning-retrieval integration for text-based clinical notes. Regarding base model selection, we focused on general-purpose foundation models rather than specialized medical LLMs (e.g., Med-PaLM 2[78]) due to access and privacy constraints. This choice ensured controlled comparisons where all methods were evaluated on identical backbones, attributing observed improvements to our proposed approaches rather than model differences. Our open-source implementation enables evaluation on medical LLMs when accessible.

Fifth, real-world validation using clinical notes from electronic health records (EHRs) is crucial to examine performance. In our study, the overall performance using in-house notes is much lower than Phenopacket-derived notes or literature-derived notes, suggesting that the quality of notes play a significant role in performance. The Phenopacket-derived notes were generated using manually annotated structured phenotype information, yet the literature-derived notes were edited by human experts for scientific publications; therefore, they can be considered as high-quality notes. Yet real clinical notes may include many noises or information irrelevant to clinical phenotypic presentations. Therefore, evaluations on publicly available notes may artificially inflate the actual performance measures.

We also note that despite the significant improvements observed with these novel techniques, the overall accuracy of LLMs in gene prioritization and rare disease diagnosis remains relatively low, highlighting the need for further advancements. Even with the best-performing approach (CoT-driven RAG), Top-10 and Top-1 accuracy for disease diagnosis and gene prioritization remains far from clinically acceptable thresholds – however, the main purpose of gene/disease prioritization is to supply candidate lists to be combined with other information, such as whole genome/exome sequencing, to speed up clinical diagnosis. Nevertheless, we believe that several key areas should be further explored to improve performance:

One major area of future optimization is retrieval quality, as RAG performance heavily depends on database coverage. Expanding retrieval sources beyond OMIM and HPO to other clinical phenotype databases such as ClinVar, Orphanet and biomedical literature, will likely improve knowledge precision and recall. Furthermore, integrating multimodal clinical data, including biochemical blood tests, electroencephalogram results, imaging scans, or even genomic sequencing, may significantly improve diagnostic accuracy. Future research can explore these directions to bridge the accuracy gap and enable reliable AI-assisted rare disease diagnosis.

In summary, with the growth of rare disease datasets with rich phenotype information (especially unstructured narratives), and the development of more sophisticated reasoning algorithms, such as graph-based reasoning[79] and neuro-symbolic reasoning[80], we expect that more advanced LLMs will successfully integrate medical knowledge while also capturing the essential phenotype-based diagnostic logic, to further improve performance of gene/disease prioritization in the future. Additionally, reinforcement learning techniques such as Group relative Policy Optimization (GRPO)[55] may further refine LLM decision-making, ensuring better alignment with phenotype-based diagnostic logic. Besides, AI-agent workflow that leverage LLM capabilities within autonomous clinical reasoning frameworks are emerging as a powerful next step. By orchestrating multiple subsystems (e.g., knowledge retrieval, context-based prompt engineering, iterative diagnosis), such agents can further streamline and automate the end-to-end diagnostic workflow[81].

## MATERIAL AND METHODS

### *Experimental Design*

In this study, we assess the performance of two LLMs (Llama 3.3-70B-Instruct and DeepSeek-R1-Distill-Llama-70B) and their RAG/CoT variations on rare genetic disorders diagnosis and gene prioritization, using four  distinct test datasets (**Table 1**): Phenopacket-derived Clinical Notes, PubMed Free Text, and In-house Clinical Notes, and Arcus clinical notes from CHOP. For both Llama 3.3-70B-Instruct and DeepSeek-R1-Distill-Llama-70B, and across each dataset, we employ four different methodologies to predict rare genetic diseases: (1) direct inference using the base prompt, (2) direct inference with the Chain of Thought (CoT) prompt, (3) direct inference with the base prompt enhanced by Retrieval-Augmented Generation (RAG), (4) the Retrieval-augmented Generation-driven Chain-of-Thought (RAG-driven CoT), and (5) Chain-of-Thought-driven Retrival-augmented Generation (CoT-driven RAG) . Both RAG-driven CoT and CoT-driven RAG integrate RAG with CoT, but they differ in their sequencing. In RAG-driven CoT, the RAG system is activated to perform searches before the model engages in structured thinking (CoT). Conversely, in CoT-driven RAG, the model first undertakes structured thinking guided by CoT and then uses the structured outputs (such as HPO terms) to query the RAG database. The graphical illustration of RAG-based methods, including (3), (4) and (5), are presented in **Figure 1**. Additionally, while we do not conduct a full-scale evaluation of OpenAI GPT-4o (Feb. 2025 version), we include one case study showcasing the power of CoT prompting.

### *Datasets*

In this paper, we assess the ability of large language models (LLMs) to diagnose rare diseases using multiple data sources. The "Phenopacket-derived Clinical Notes" (**Table 1**) dataset comprises clinical notes derived from structured Phenopacket-store data[57]. Originally, the Phenopacket-store data included details such as sex and HPO IDs, which we used to prompt GPT-4 to generate extensive clinical notes. These unstructured clinical notes from the Phenopacket-store are also publicly accessible on our GitHub page. The prompt used for generating synthetic clinical notes is presented in **Table S1**.

To conserve computational time and resources and ensure a more balanced dataset, we limit our selection to a maximum of two patients per disease/gene. Specifically, if a disease/gene has more than two associated patients, we randomly choose two patients for inclusion.

In addition to Phenopacket, we utilized another dataset, referred to as "PubMed Free Text" in **Table 1**, which includes 255 free text paragraphs detailing patient demographics and HPO terms. This dataset was sourced from publications and is publicly accessible on the GitHub page of the previous study[38]. We also gathered in-house clinical notesfrom the Children's Hospital of Philadelphia (CHOP), known as "In-house Clinical Notes" in **Table 1**. These notes, manually selected for their clear disease/gene labels and rich patient information, are not made public due to regulatory requirements. **Table 1** provides an overview of datasets used for evaluation in this study. We further included an Arcus dataset composed of noisy, free-form clinical notes, analogous in nature to our in-house cohort. Arcus is an information platform developed to support researchers providing access to CHOP clinical and research data and data analysis tools in a customized computing space. The Arcus dataset contains 868 cases for disease diagnosis and 702 cases for gene prioritization for whom positive diagnosis are already known from outside genetic diagnostic labs.

To contextualize the clinical and genetic diversity represented in our evaluation, we performed a coarse functional categorization of representative diseases and genes from the PubMed and in-house cohorts. The resulting categories (e.g., neurodevelopmental/neurological, metabolic/mitochondrial, craniofacial/developmental, and others) and representative examples are summarized in **Table S4**, illustrating that both cohorts span a broad range of diagnostic domains rather than a single disease class.

### *Ethics Statement*

This study was approved by the Institutional Review Board (IRB) of Children's Hospital of Philadelphia (CHOP). All research activities involving in-house clinical notes were conducted in strict compliance with the Health Insurance Portability and Accountability Act (HIPAA) and institutional data governance policies. To ensure patient privacy protection, all computational analyses of protected health information were performed within approved secure computing environment, including the institutional computing cluster or the Arcus data platform. Patient identifiers were not included in any outputs or shared materials. The synthetic datasets generated from Phenopacket-store and publicly available data sources do not contain protected health information and are made available for research purposes, so that others can reproduce our analytical protocol and results.

### *Evaluation Metric*

In the context of phenotype-driven rare disease diagnosis and gene prioritization, accuracy is the most crucial for assessing how effectively a system ranks the correct disease or causative gene from phenotypic information from all candidates.

To mirror the practical realities of clinical decision-making, the Top-1 and Top-10 accuracies are most commonly used in the evaluation. Top-1 accuracy denotes the proportion of test cases in which the ground truth appears as the highest-ranked result, while Top-10 denotes the proportion of test cases in which the ground truth appears anywhere among the top ten predictions.

By prioritizing these metrics, evaluations can closely approximate routine medical practice, in which clinicians typically consult only a small number of top-ranked results. Broader rankings such as Top-50 and Top-100, are more practical in scenarios where the ranked list need to be combined with other information, such as whole genome/exome sequencing data. Our study evaluated Top-1 and Top-10 accuracy because we need to reduce the computational cost for LLMs, and because we do not have additional whole genome/exome sequencing data to further trim down the candidates.

### *Prompting Strategies*

To fully explore the capabilities of large language models (LLMs), we examine the effects of various prompts on gene prioritization and rare disease diagnosis tasks. A prompt is an instruction given to the LLM to steer its response[82]. Considering the text-completion functionality of these models, the quality of prompts likely plays a critical role in influencing outcomes[83]. We also speculate that the nature of the input, whether structured or noisy clinical texts, might be better suited to different types of prompts. The noisier the input, the greater the need for human judgment or guidance in crafting these prompts. When the input to the LLMs consists of unstructured clinical texts, we aim to evaluate the model's ability to engage in structured reasoning before generating an output, using one-shot learning through CoT prompts. **Table 2** and **Table S1** summarize all the prompts considered in this study.

### *Chain of Thought (CoT) Prompting*

We utilize Chain-of-Thought (CoT) prompting method that guides the model through a structured, five-step process for interpreting clinical notes. In the disease diagnosis task, the model (1) first extract and classifies HPO terms, organizing phenotypic features (e.g., “genitourinary system”, “nervous system”) for clarity. (2) Next, it assesses demographic impact, recognizing how age, gender, and ethnicity might shape disease likelihood. (3) The third step categorizes possible conditions into broad disease groups, establishing a preliminary diagnostic scope. The model then narrows down candidate diseases by ruling out inconsistencies (e.g., inheritance patterns or phenotype severity), culminating in a ranked list of exactly ten diseases. This final output is strictly formatted, ensuring efficient clinical review.

An equivalent five-step approach underlies our gene prioritization workflow. Again, the model extracts and classifies HPO terms, followed by assessment of demographic factors. It then maps these phenotypic features to relevant gene-disease associations, referencing known literature and variants. And finally, the model refines the gene list based on inheritance patterns and the patient's representation, by mandating clear reasoning at each phase and prescribing uniform output requirement (e.g., "Exactly 10" labels). Note that the CoT prompts for gene prioritization and rare disease diagnosis can be found in **Table 2** and **Table S1** respectively.

***Retrieval Augmented Generation (RAG)***

In this study, we use two data sources to develop our Retrieval-Augmented Generation (RAG) system: the HPO database[84] and OMIM texts[65] . The curated HPO database provides structured data, such as age, gender, disease name, gene type, HPO terms grouped by organ system and a textual description describing a rare syndrome, for example, a typical HPO record might list a disease name like "Sotos syndrome". Indicate "NSD1/SETD2/APC2" for the gene type, note that the typical age (e.g. 1-5 years old) and gender (e.g. non-sex specific), include a series HPO terms grouped into categories based on affected organ systems (e.g., Ear: Posteriorly rotated ears; Endocrine system: Hypothyroidism), along with a textual definition summarizing the syndrome. The OMIM texts, which are sourced from the OMIM website, consist of human-curated summaries of disease case studies drawn from various publications. Therefore, our RAG system incorporates both structured and unstructured databases, offering a robust platform capable of managing both structured and non-structured inputs. The complete process of accessing the HPO and OMIM databases is shown in the *Supplementary Materials*.

Our pipeline is built using LangChain[85,86] and incorporates a two-stage retrieval process. (1) An embedding-based retriever utilizes NeuML/pubmedbert-base-embeddings[29], a sentence transformer[87] fine-tuned on PubMed title-abstract pairs to better capture biomedical semantics, and FAISS[88] for efficient dense retrieval. Retrieved documents are processed using recursive chunking (chunk size: 512 tokens) to ensure semantic coherence before passing them to the LLM backbone for response generation. We retrieve the top 3 most relevant documents at this stage. (2) A reranker (cobert-ir/colbertv2.0[89,90]) refines the ranking via late interaction scoring, selecting the top 1 document as context. Based on document length distribution shown in **Table 1** and **Figure 4**, we adopt a context window of 2048 tokens for Phenopacket-derived clinical notes and PubMed Free-text, while using 5120 tokens for in-house clinical notes to accommodate longer texts efficiently.

***Retrieval-augmented Generation (RAG)-driven Chain-of-Thought (CoT)***

In the RAG-driven CoT pipeline, retrieval occurs first. Rather than prompting the model to engage in CoT immediately, the system begins by tapping into our RAG framework (as detailed in the ***Materials and Methods*** section on RAG) to gather potentially relevant documents from the HPO and OMIM databases. Once these external resources have been identified, the model applies its previously defined CoT workflow (see CoT description) to synthesize the retrieved information with patients' clinical data. This approach is particularly effective in high-quality contexts, where early retrieval can anchor the subsequent reasoning steps in domain-specific evidence.

### *Chain-of-Thought (CoT)-driven Retrieval-augmented Generation (RAG)*

CoT-driven RAG reverses the sequence of operations compared to RAG-driven CoT. The model first executes its five-step chain-of-thought on the clinical information-classifying HPO terms, assessing demographics, and delineating initial hypothesis, before tapping into the RAG system. The refined query that emerges from these structured reasoning steps allows for more targeted retrieval, which is often advantageous when processing lengthy or noisy clinical notes. After retrieval, the model integrates the relevant passages into its final diagnosis or gene prioritization output, ensuring that extraneous or irrelevant materials have minimal impact on the concluding predictions.

In both hybrid methods, the structured CoT prompts and the RAG system remain fundamentally the same. The key distinction is the order in which external retrieval and chain-of-thought reasoning are invoked. Consequently, users may select the pipeline most appropriate for their dataset, RAG-driven CoT can enrich early reasoning with domain context, whereas CoT-driven RAG can refine queries to avoid extraneous or unrelated content.

## ACKNOWLEDGEMENTS

We thank the CHOP Arcus team for provisioning computing infrastructure and virtual servers for data analysis on in-house clinical notes, and thank IDDRC Biostatistics and Data Science core (HD105354) for technical consultation. This study was supported by NIH/NHGRI grant HG013031 and the CHOP Research Institute.

## DECLARATIONS

### *Availability of data and materials*

All the datasets except in-house clinical notes from CHOP are publicly available at the following GitHub repository: https://github.com/WGLab/CoT-RAG-LLM-Gene-Prioritization-Disease-Diagnosis

### *Code availability*

All the code can be accessed publicly on the following GitHub repository: https://github.com/WGLab/CoT-RAG-LLM-Gene-Prioritization-Disease-Diagnosis

### *Competing interests*

The authors declare no competing interests.

## FIGURES

**Figure 1. Illustration of workflow of different methods in this study**. **A**: Base Prompt + CoT, **B**: Base Prompt + RAG, **C**: CoT-driven RAG, and **D**: RAG-driven CoT.

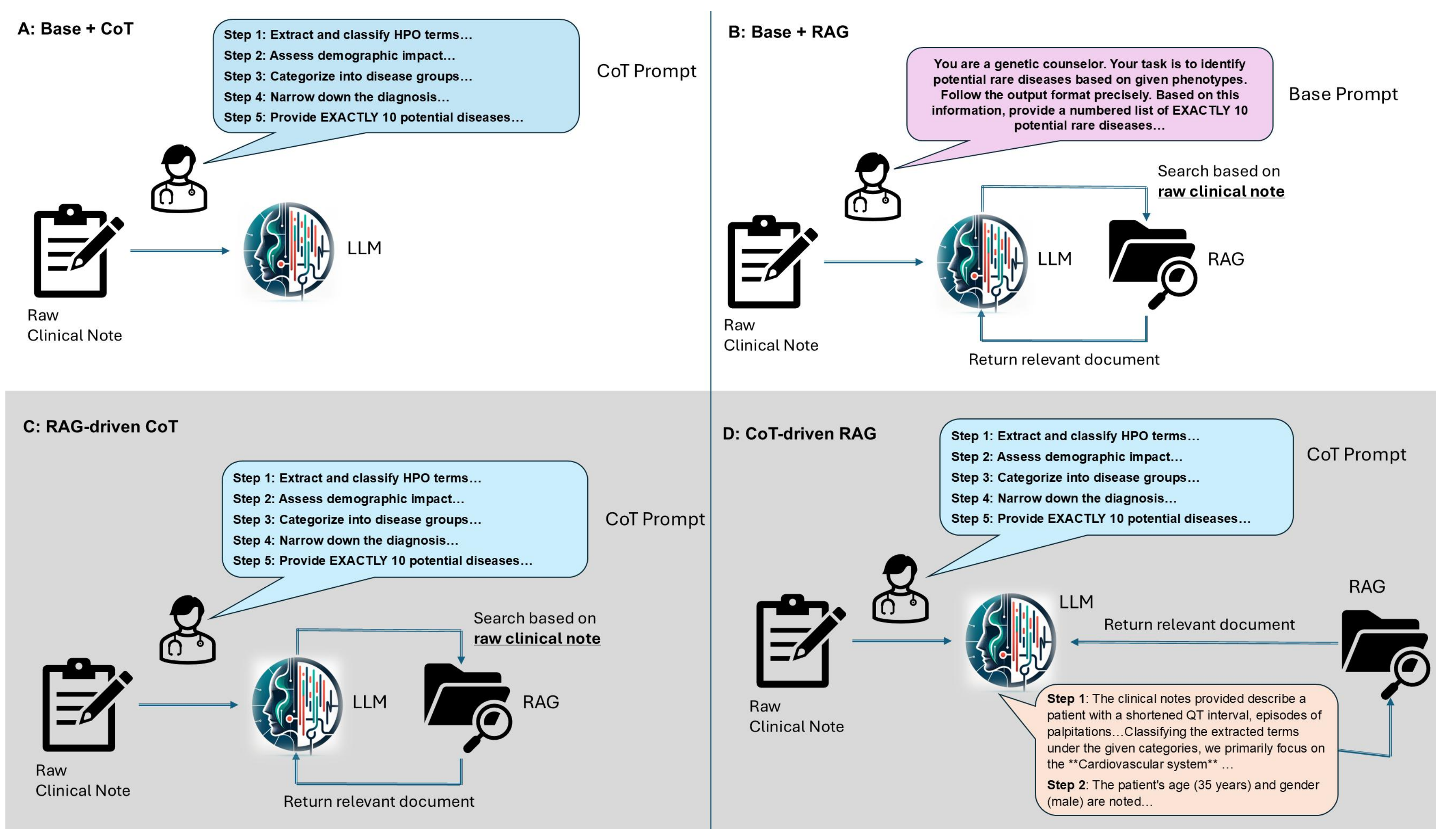

**Figure 2. The Top-10 Accuracy For Llama 3.3-70B-Instruct and DeepSeek-R1-Distill-Llama-70B in Gene Prioritization and Rare Diseases Diagnosis.** All the corresponding numerical results (including Top-1 accuracy) are documented in **Table S2**.

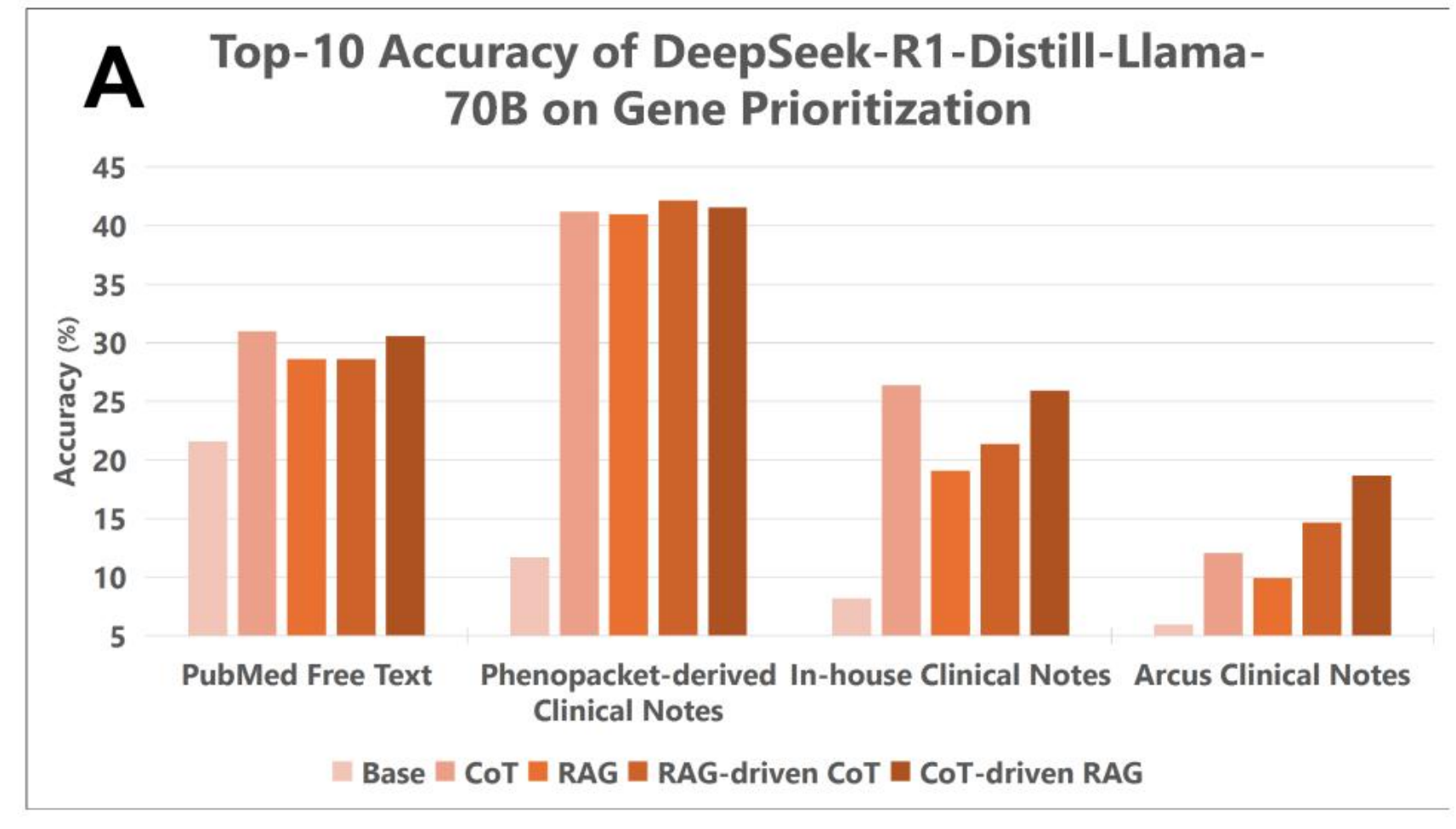

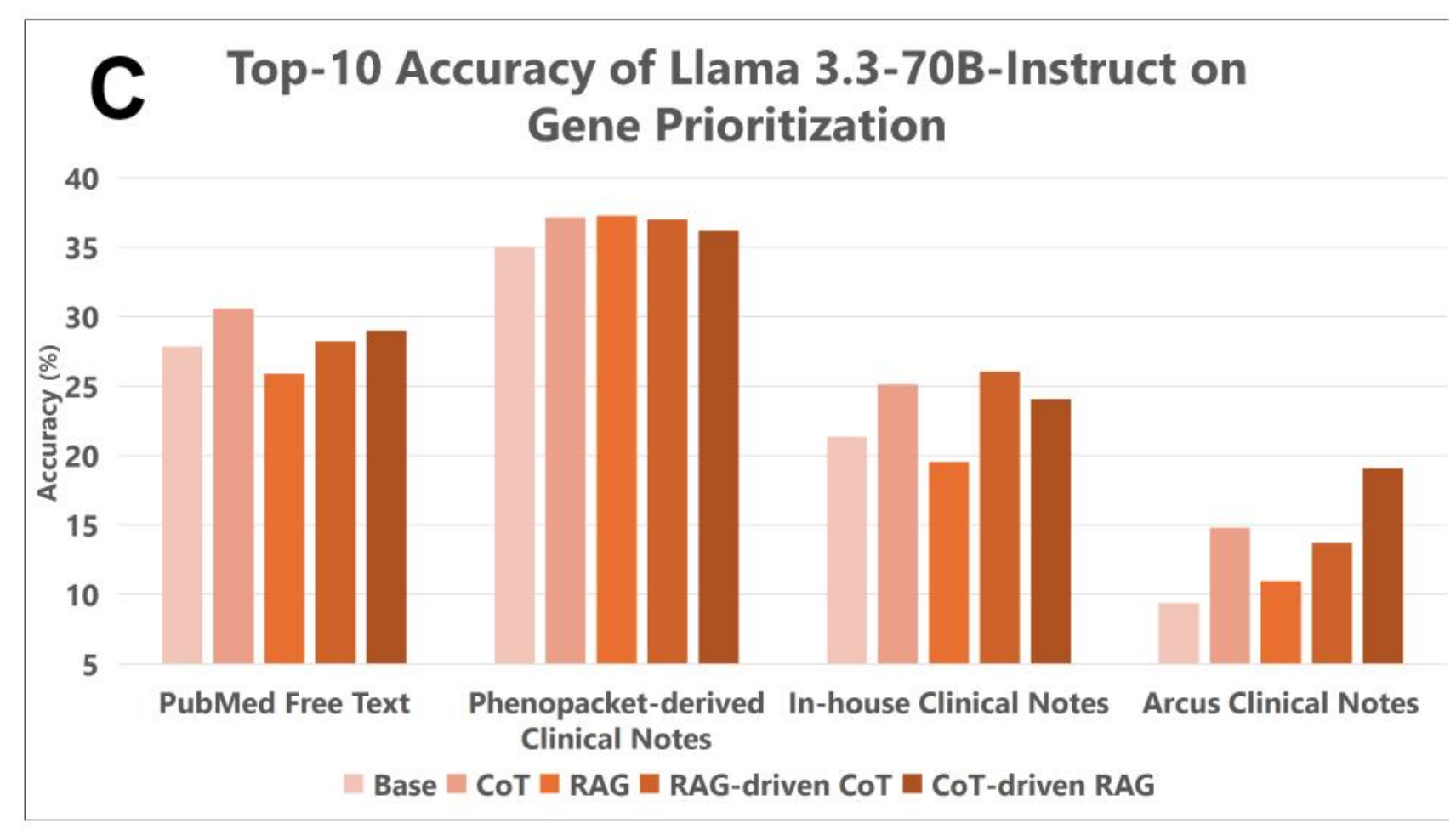

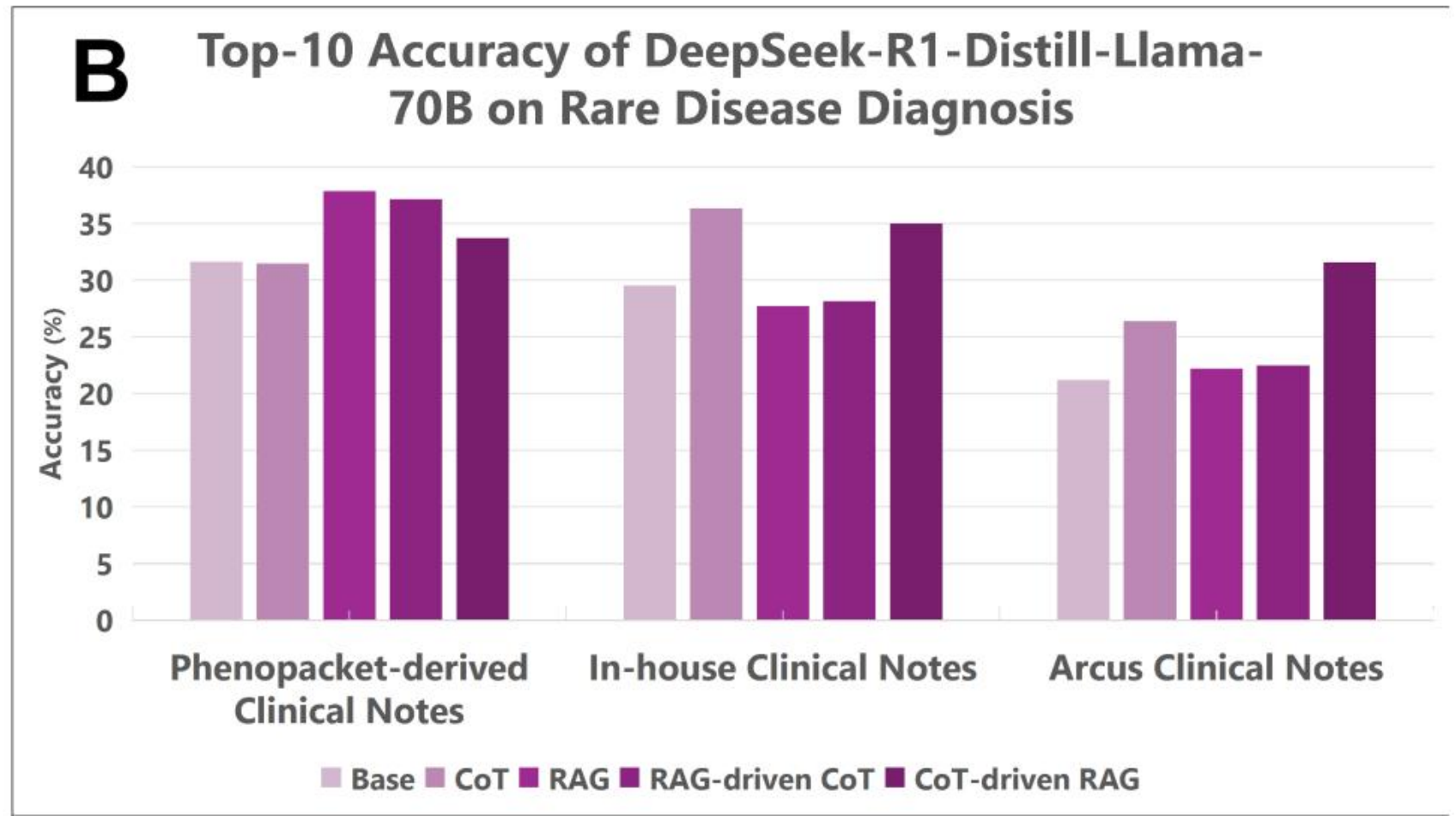

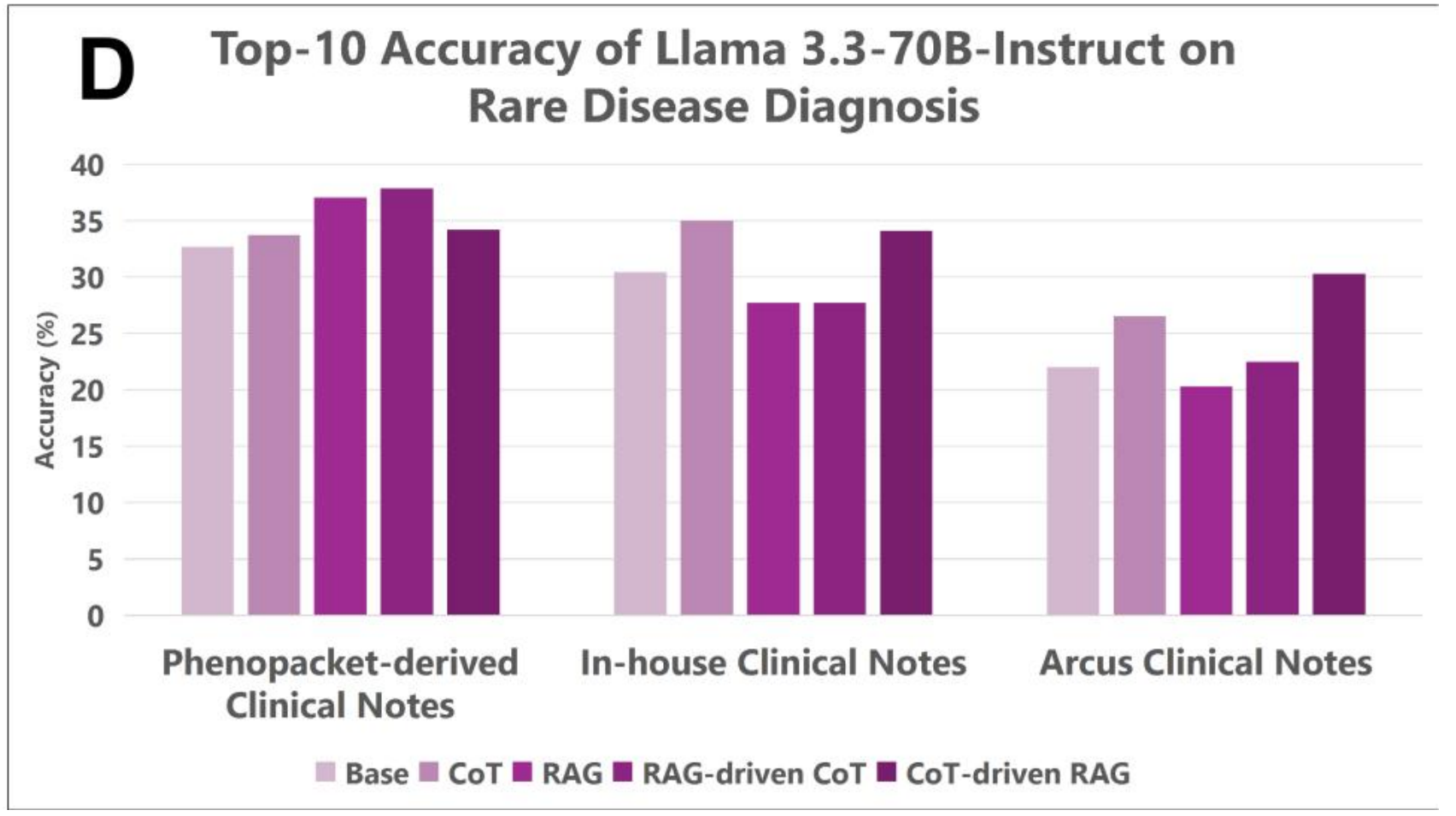

**Figure 3. PacMAP Projection of  Embeddings and Length Distributions of RAG Knowledge Databases.** (a) PaCMAP projection revealing three distinct clusters: HPO Database documents (I, purple), mixed OMIM and HPO documents (II, teal), and OMIM Database documents (III, orange). (b) PacMAP of documents embeddings from OMIM (green) and HPO (yellow) with RAG-driven CoT query vectors (Phenopacket-derived clinical notes: diamonds, PubMed Free-text: squares, in-house clinical notes: crosses) distributed across multiple clusters. (c) Same embeddings with CoT-driven RAG query vectors. (d) Distribution of document length in Knowledge Base. This histogram shows the distribution of document lengths (in tokens) across our knowledge base containing 64,068 documents. The light blue bars represent frequency counts, while the red curve indicates the kernel density estimate. Vertical dashed lines mark the mean (289.2 tokens, orange) and median (285.0 tokens, green) values. (e) Document length distribution of OMIM Database containing 48,136 documents. (f) Document length distribution of HPO Database containing 15,762 documents.

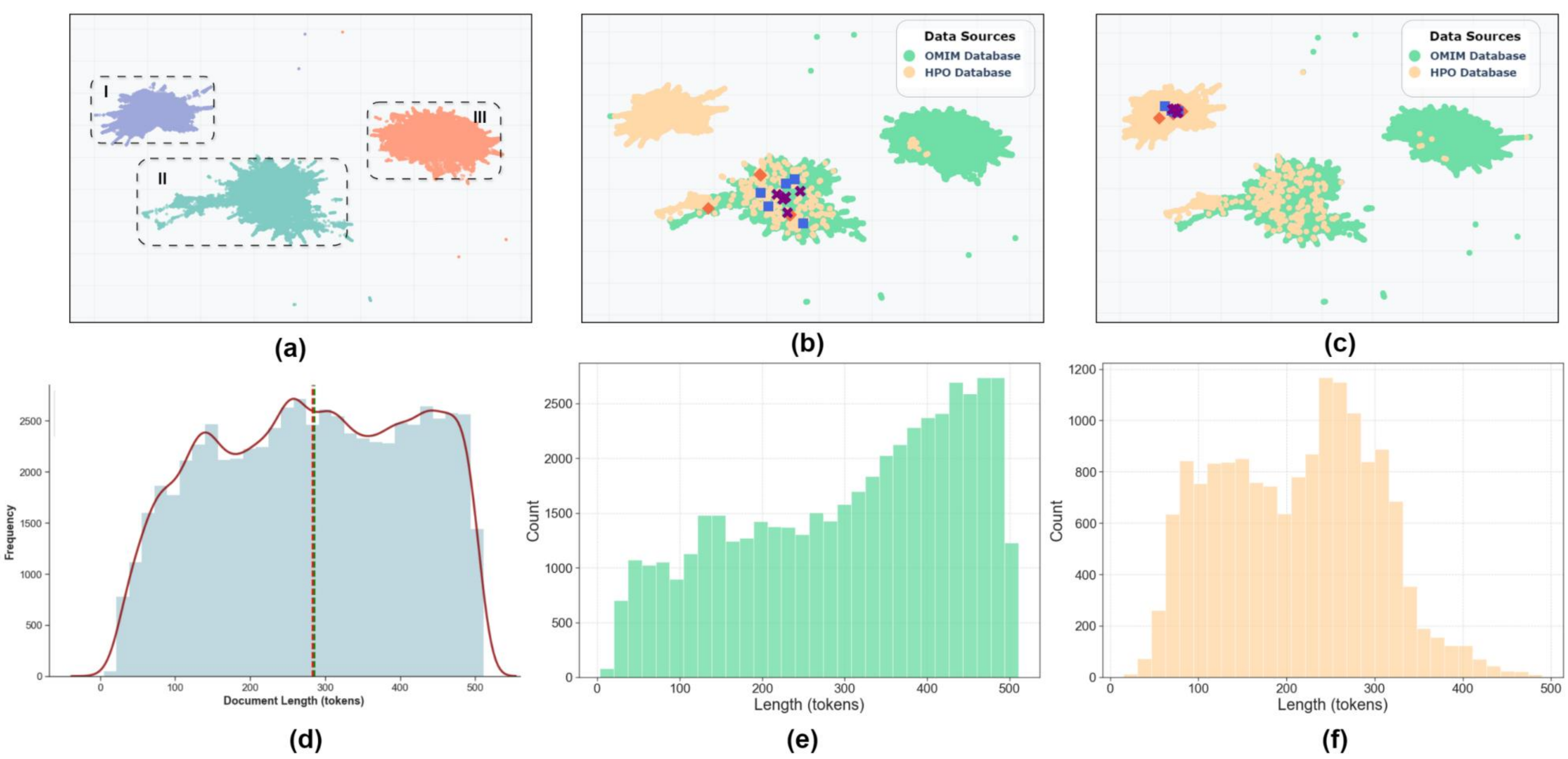

## TABLES

**Table 1: Overview of datasets used for evaluation in this study.** The column labeled "Public" (Yes/No) specifies whether the listed dataset is available to the public.

| Overview of Datasets Used for Evaluation | | | | | | |
|---|---|---|---|---|---|---|
| Data Source | Public | Total # of samples | # of samples used for evaluation for each task | | Average # of words (±std) | Total # of diseases/genes covered |
| | | | Disease Diagnosis | Gene Prioritization | | |
| Phenopacket-derived clinical notes | Yes | 5980 | 845 | 845 | 298.20±49.90 | 468 / 418 |
| PubMed Free Text | Yes | 255 | 0 | 255 | 198.51±126.96 | 0 / 95 |
| In-house Clinical Notes | No | 220 | 220 | 220 | 1606.35±789.71 | 114 / 146 |
| Arcus Dataset | No | 1088 | 868 | 702 | 976,66±499.65 | 610 / 388 |

**Table 2: Summary of prompts for gene prioritization used in this study.** The prompts (both base prompt and Chain of Thought) used for rare disease diagnosis closely resemble those employed for gene prioritization, so provide these details in the Supplementary Materials (**Table S1**).

| Summary of Prompts Used in This Study | | |
|---|---|---|
| Prompt Template | Task | Prompt Content |
| Base Prompt | Gene Prioritization | You are a genetic counselor. Your task is to identify potential genes associated with the given phenotypes. Follow the output format precisely. **'[clinical_note]'** Based on this information, provide a numbered list of EXACTLY 10 potential genes.\n\nUse EXACTLY this format:\n\nPOTENTIAL_GENES:\n1. 'Gene1'\n2. 'Gene2'\n3. 'Gene3'\n4. 'Gene4'\n5. 'Gene5'\n6. 'Gene6'\n7. 'Gene7'\n8. 'Gene8'\n9. 'Gene9'\n10. 'Gene10'\n\nEnsure all gene names are in single quotes, and there are exactly 10 in the list. Do not deviate from this format or add any explanations. |
| Chain-of-Thought (CoT) Prompt | Gene Prioritization | **System Message:** You are a geneticist specializing in gene-disease associations. Your task is to reason step-by-step based on the given clinical notes and prioritize EXACTLY 10 genes most likely associated with the described phenotypes. Use the reasoning process outlined below.<br><br>User Message:<br><br>'[clinical_note]'<br><br>Based on the provided clinical notes, follow this reasoning process:<br><br>1. **Extract and classify HPO terms**<br><br>###Classify the extracted terms under the following categories:<br><br>**Genitourinary system**;**Cellular phenotype**;**Blood and blood-forming tissues**;**Head and neck**;**Limbs;**Metabolism/homeostasis;**Prenatal development or birth;**Breast;**Cardiovascular system;**Digestive system;**Ear;**Endocrine system;**Eye;**Immune |
| | | system;**Integument;**Musculoskeletal system;**Nervous system;**Respiratory system;**Thoracic cavity;**Voice;**Constitutional symptoms;**Growth abnormality;**Neoplasm**.<br><br>2. **Assess demographic impact**: How do the patient's **age, gender, and ethnicity** impact the likelihood of gene involvement?<br><br>3. **Map to relevant gene-disease associations**: Based on the extracted HPO terms, retrieve relevant genes known to be associated with these phenotypes.<br><br>4. **Refine based on inheritance patterns and variant evidence**: Narrow down genes based on mode of inheritance, functional impact, and known pathogenic variants.<br><br>5. **Prioritize the top 10 genes**: Finalize a ranked list of EXACTLY 10 genes that are most likely associated with the described phenotypes.<br><br>Output your reasoning for each step clearly, followed by the final gene list. Use the following format:<br><br>REASONING:<br><br>Step 1: [Extract and classify HPO terms]<br><br>Step 2: [Assess demographic impact]<br><br>Step 3: [Map to relevant gene-disease associations]<br><br>Step 4: [Refine based on inheritance patterns and variant evidence]<br><br>Step 5: [Prioritize the top 10 genes]<br><br>POTENTIAL_GENES:\n1. 'Gene1'\n2. 'Gene2'\n3. 'Gene3'\n4. 'Gene4'\n5. 'Gene5'\n6. 'Gene6'\n7. 'Gene7'\n8. 'Gene8'\n9. 'Gene9'\n10. 'Gene10'\n\nEnsure all gene names are in single quotes, and there are exactly 10 in the list. Do not deviate from this format or add any explanations. |

**Table 3.** Comparison of causal gene ranking across different inference strategies in two case studies.

| Case | Causal gene | Gene rank by models | Clinical Notes |
|---|---|---|---|
| 1 | ACTA1 | 6/base, 1/RAG, 1/RAG-driven CoT, 1/CoT-driven RAG | "The patient, a female, presents with a distinctive clinical profile primarily characterized by hypotonia and areflexia, suggestive of neuromuscular involvement. The hypotonia is manifesting as a reduced muscle tone, observable throughout her musculature, paired with a lack of reflexive responses during neurological examination. She also exhibits proximal muscle weakness, notable in her larger muscle groups which suggests that her muscular impairment is more significant towards the axis of the body. This is coupled with weakness of her facial musculature, evident in a bland facial expression and potential functional challenges with facial activities.<br><br>Her oral examination reveals a high palate, a structural anomaly that can be associated with skeletal muscle conditions, potentially affecting feeding and speech. Despite these findings, intellectual development remains within normal limits, excluding intellectual disability as a contributing factor to her presentation. There is an absence of distal muscle weakness, with muscle strength in her hands and feet being relatively preserved, and no evidence of ophthalmoplegia that would indicate ocular muscle involvement.<br><br>Motor development has been delayed, as noted in her early childhood records, an element that aligns with her general muscle weakness. Musculoskeletal examination also reveals an ankle flexion contracture, suggesting chronic muscle shortening possibly due to sustained weakness and imbalance. Importantly, there is no history of drooling or bulbar palsy, indicating that the motor issues have not significantly impacted her bulbar function.<br><br>The overall symptomatology matches a clinical entity characterized by muscle weakness starting in childhood, which has been genetically corroborated. The patient's genetic testing revealed a heterozygous variant, consistent with a diagnosis linked to congenital muscle disorder, established through causative genomic interpretation. This condition generally follows an autosomal dominant pattern and is known to present with the described clinical features. The exclusion of significant intellectual impairment and the absence of distal muscle and eye muscle involvement further refine the diagnostic picture in this case." |
| 2 | ZIC3 | 4/base,<br>1/CoT | “The patient, a male, presents with several significant congenital cardiac anomalies and splenic abnormalities. From birth, he has exhibited a single ventricle heart defect, a complex condition where one of the ventricles is underdeveloped or absent, affecting normal circulatory function. Alongside this, he is affected by polysplenia, a condition characterized by the presence of multiple small accessory spleens instead of a single, normal-sized spleen. Fortunately, asplenia, the complete absence of splenic tissue, has been ruled out in this patient.<br><br>Further complicating his clinical picture is the congenital presence of dextrotransposition of the great arteries (d-TGA), where the positions of the main arteries, namely the aorta and the pulmonary artery, are switched, resulting in significant circulatory issues if not corrected. Additionally, the patient has pulmonary valve atresia, a defect where the pulmonary valve does not form properly, impeding blood flow from the right ventricle to the lungs, which further impacts oxygenation of the blood.<br><br>Despite these multiple cardiac anomalies, the patient does not exhibit mitral atresia, suggesting that the mitral valve, which regulates blood flow between the left atrium and left ventricle, is present and functional to some extent.<br><br>This comprehensive clinical presentation is consistent with a diagnosis that has been confirmed as an underlying genetic condition, which typically emerges from birth. The diagnosis has been further substantiated at the genetic level, confirming the presence of a causative mutation responsible for the array of presented phenotypes. This condition reflects a complex interplay of anomalies that indicate a significant impact on the patient's cardiac function and splenic morphology from an early stage of life. The clinical course necessitates a multi-disciplinary approach to address these congenital issues, with potential surgical and medical interventions to manage symptoms and improve quality of life.” |

# Supplementary Materials

**Table S1: Summary of prompting strategies for generating synthetic datasets from Phenopacket-Store  and rare diseases diagnosis.** The prompts for gene prioritization are documented in **Table 2.**

| **Prompts for Rare Diseases Diagnosis** | | |
|---|---|---|
| Prompt Template | Task | Prompt Content |
| Prompt for Synthetic Data Generation | Construction of Noisy Clinical Notes from Structured Phenopacket Data | You are an experienced clinician writing realistic and professional clinical notes. Given a structured JSON dataset containing a patient’s demographic details and phenotypic features, generate a detailed and natural clinical note in a free-flowing narrative format, similar to real-world documentation. Ensure the note reads as though written by a real clinician, maintaining a natural clinical tone. Avoid structured headings or bullet points. Instead, present the information as a cohesive, flowing clinical note. Describe the patient’s symptoms, physical features, and clinical concerns based on the given phenotypic features. Do not mention gene names, variants, or explicit disease names. Focus on describing the clinical presentation based on the given data. |
| Base Prompt | Rare Disease Diagnosis | You are a genetic counselor. Your task is to identify potential rare diseases based on given phenotypes. Follow the output format precisely. Based on this information, provide a numbered list of EXACTLY 10 potential rare diseases.\n\nUse EXACTLY this format:\n\nPOTENTIAL_DISEASES:\n1. 'Disease1'\n2. 'Disease2'\n3. 'Disease3'\n4. 'Disease4'\n5. 'Disease5'\n6. 'Disease6'\n7. 'Disease7'\n8. 'Disease8'\n9. 'Disease9'\n10. 'Disease10'\n\nEnsure all disease names are in single quotes, and there are exactly 10 in the list. Do not deviate from this format or add any explanations. |

| CoT Prompt | Rare Disease Diagnosis | Step 1: Extract and classify HPO terms<br><br>Extract key **phenotypic features (HPO terms)** from the following clinical note and classify them according to the **Human Phenotype Ontology (HPO) system**:<br><br>Classify the extracted terms under the following categories:<br>- Genitourinary system<br>- Cellular phenotype<br>- Blood and blood-forming tissues<br>- Head and neck<br>- Limbs<br>- Metabolism/homeostasis<br>- Prenatal development or birth<br>- Breast<br>- Cardiovascular system<br>- Digestive system<br>- Ear |
|---|---|---|
| | | - Endocrine system<br>- Eye<br>- Immune system<br>- Integument<br>- Musculoskeletal system<br>- Nervous system<br>- Respiratory system<br>- Thoracic cavity<br>- Voice<br>- Constitutional symptoms<br>- Growth abnormality<br>- Neoplasm<br>Now, classify the **HPO terms** for the following patient:<br>“[clinical_note]”<br>) |

| | | Step 2: Assess demographic impact<br><br>How do the patient's **age, gender, and ethnicity** impact disease likelihood?<br><br><br>Step 3: Categorize into disease groups<br><br>Based on the extracted HPO terms, which broad disease categories are most relevant?<br><br><br>Step 4: Narrow down the diagnosis<br><br>Narrow down the diseases by eliminating less likely options based on phenotype severity, inheritance patterns, and clinical features.<br><br><br>Step 5: Provide EXACTLY 10 potential diseases<br><br>Finalize a ranked list of EXACTLY 10 diseases that could explain the patient's presentation.<br><br>POTENTIAL DISEASES:<br><br>1. 'Disease1'<br><br>2. 'Disease2'<br><br>3. 'Disease3'<br><br>4. 'Disease4'<br><br>5. 'Disease5' |
|---|---|---|
| | | 6. 'Disease6'<br><br>7. 'Disease7'<br><br>8. 'Disease8'<br><br>9. 'Disease9'<br><br>10. 'Disease10'<br><br>Ensure all disease names are in single quotes, and there are exactly 10 in the list. Do not add any additional explanations or deviate from the specified format. |

**Table S2: Accuracy of top-10 and top-1 prediction by various methods**. Note that the hallucination metric (E1) is consistently 100% across all cases, which is why it is omitted from the table. Numbers in boldface represent the highest performance within each specific testing cohort. The graphical illustration of this table is **Figure 2**.

| Results by Llama 3.3-70B-Instruct | | | | | | | | | | | | |
|---|---|---|---|---|---|---|---|---|---|---|---|---|
| Dataset/ Add-on Technique | Task Label | Size | Top-10 Accuracy | | | | | Top-1 Accuracy | | | | |
| | | | Base | CoT | RAG | RAG-driven CoT | CoT-driven RAG | Base | CoT | RAG | RAG-driven CoT | CoT-driven RAG |
| PubMed Free Text | Gene | 255 | 27.84 | **30.58** | 25.88 | 28.24 | 29.02 | 16.47 | **16.47** | 9.80 | 12.16 | 16.08 |
| Phenopacket-derived Clinical Notes | Disease | 845 | 32.68 | 33.72 | 37.05 | **37.87** | 34.20 | 20.74 | 22.13 | 23.31 | **24.97** | 22.37 |
| Phenopacket-derived Clinical Notes | Gene | 845 | 35.03 | 37.16 | **37.28** | 37.01 | 36.21 | 18.34 | 19.17 | 17.87 | **21.11** | 19.64 |
| In-house Clinical Notes | Disease | 220 | 30.45 | **35.00** | 27.73 | 27.73 | 34.09 | 11.82 | 16.82 | 11.82 | 16.36 | **17.27** |

| Arcus Clinical Notes | Disease | 868 | 22.00 | 26.52 | 20.28 | 22.47 | **30.30** | 8.18 | 10.32 | 9.10 | 9.33 | **11.41** |
|---|---|---|---|---|---|---|---|---|---|---|---|---|
| In-house Clinical Notes | Gene | 220 | 21.36 | 25 .11 | 19.55 | **26.03** | 24.09 | 10.00 | 11.87 | 5.91 | 10.95 | **12.27** |
| Arcus Clinical Notes | Gene | 702 | 9.40 | 14.81 | 10.97 | 13.68 | **19.09** | 4.98 | 7.55 | 5.69 | 6.13 | **12.68** |
| Results by DeepSeek-R1-Distill-Llama-70B | | | | | | | | | | | | |
| Dataset/ Add-on Technique | Task Label | Size | Top-10 Accuracy (E3) | | | | | Top-1 Accuracy (E2) | | | | |
| | | | Base | CoT | RAG | RAG-driven CoT | CoT-driven RAG | Base | CoT | RAG | RAG-driven CoT | CoT-driven RAG |
| PubMed Free Text | Gene | 255 | 21.57 | **30.98** | 28.63 | 28.63 | 30.58 | 7.06 | 18.82 | 16.47 | 17.65 | **20.00** |
| Phenopacket-derived Clinical Notes | Disease | 845 | 31.65 | 31.51 | **37.87** | 37.15 | 33.73 | 20.37 | 22.88 | 26.86 | **28.19** | 22.95 |
| Phenopacket-derived Clinical Notes | Gene | 845 | 11.72 | 41.18 | 40.95 | **42.13** | 41.54 | 5.68 | 23.07 | 21.18 | **23.78** | 23.20 |
| In-house Clinical Notes | Disease | 220 | 29.55 | **36.36** | 27.73 | 28.18 | 35.00 | 15.91 | 17.27 | 12.27 | 16.82 | **17.72** |
| Arcus Clinical Notes | Disease | 868 | 21.19 | 26.38 | 22.23 | 22.47 | **31.57** | 11.18 | 13.94 | 10.71 | 12.55 | **14.86** |
| In-house Clinical Notes | Gene | 220 | 8.18 | 26.36 | 19.09 | 21.36 | **25.91** | 6.36 | 13.64 | 10.00 | 10.91 | **14.09** |
| Arcus Clinical Notes | Gene | 702 | 5.98 | 12.05 | 9.93 | 14.67 | **18.67** | 4.13 | 6.75 | 4.34 | 11.25 | **12.25** |

**Table S3. Results of two-proportion z-tests (binomial approximation) applied to Top-1 accuracy across different datasets and methods.** Reported p-values are based on sample sizes corresponding to each dataset. Statistical significance was assessed at the conventional threshold of $p < 0.05$.

| Dataset | RAG-driven CoT vs CoT | RAG-driven CoT vs RAG | CoT-driven RAG vs CoT | CoT-driven RAG vs RAG |
|---|---|---|---|---|
| **PubMed Free Text (Gene, Llama3.3)** | $p = 0.164$ | $p = 0.395$ | $p = 0.905$ | **$p = 0.035$** |
| **Phenopacket Clinical Notes (Disease, Llama3.3)** | $p = 0.169$ | $p = 0.426$ | $p = 0.907$ | $p = 0.643$ |
| **Phenopacket Clinical Notes (Gene, Llama3.3)** | $p = 0.332$ | **$p = 0.097$** | $p = 0.806$ | $p = 0.350$ |
| **In-house Clinical Notes (Disease, Llama3.3)** | $p = 0.898$ | $p = 0.171$ | $p = 0.899$ | $p = 0.105$ |
| **In-house Clinical Notes (Gene, Llama3.3)** | $p = 0.764$ | $p = 0.059$ | $p = 0.884$ | **$p = 0.020$** |
| **PubMed Free Text (Gene, DeepSeek)** | $p = 0.731$ | $p = 0.724$ | $p = 0.737$ | $p = 0.302$ |
| **Phenopacket Clinical Notes (Disease, DeepSeek)** | **$p = 0.012$** | $p = 0.549$ | $p = 0.954$ | $p = 0.063$ |
| **Phenopacket Clinical Notes (Gene, DeepSeek)** | p = 0.730 | p = 0.200 | p = 0.954 | p = 0.320 |
| **In-house Clinical Notes (Disease, DeepSeek)** | p = 0.899 | p = 0.176 | p = 0.900 | p = 0.109 |
| **In-house Clinical Notes (Gene, DeepSeek)** | p = 0.383 | p = 0.755 | p = 0.890 | p = 0.187 |

**Table S4: Functional Categorization of Diseases and Genes from PubMed, In-house, and Arcus Clinical Datasets.** Comprehensive functional categorization of diseases and genes across three independent data sources: PubMed literature-based narratives (255 cases, 74 unique genes), in-house clinical notes (220 cases, 85 unique diseases, 103 unique genes), and Arcus clinical dataset (387 unique genes, 560 unique diseases). The table demonstrates extensive coverage across all major genetic disease categories, with complementary representation from multiple clinical contexts.

| Functional Category | In-house Diseases | Arcus Diseases | PubMed Genes | In-house Genes | Arcus Genes |
|---|---|---|---|---|---|
| Craniofacial / Developmental | Cornelia de Lange, Kabuki, Rubinstein–Taybi, Coffin–Siris, Wiedemann–Steiner, Kleefstra, Koolen–de Vries, Pitt–Hopkins, Bainbridge–Ropers, White–Sutton, Floating–Harbor, CHARGE, Sotos, Tatton–Brown–Rahman | 15q14 microdeletion syndrome; 22q11.2 deletion syndrome; 9p deletion syndrome; Cornelia de lange syndrome; Kabuki syndrome; Kleefstra; Malan syndrome; Wiedemann steiner syndrome | SALL4, TBX4, SOX10, PAX6, BRPF1, BCOR, MED12, TCOF1, EYA1, NSD1, SMCHD1, ARX, ASXL3, ATN1, ZMYND11, FGD1, FOXP2, ANKRD11, NR2F1, SOX9 | ZNF462, CHD7, MED12L, KMT2A, ASXL3, SRCAP, RBM10, GNAS, STAT1, RERE, TCF4, ANKRD11, HDAC8, BRCA2, FLNA, GATA3, KIT, SMC1A, AMH, SMAD4, KDM6A, RAD21, LIG3, PTPN11, TP63, NSD1, JARID2, POMT2, KMT2D, EHMT1, CREBBP, TBX3, NFIX, FOXC1, ARID1A, PAX3 | ANKRD11, ARID1A, ARID1B, ASXL1, ASXL3, BRPF1, CHD3, CHD4, CHD7, CHD8, CREBBP, EHMT1, EYA1, FGFR2, FGFR3, FOXP1, GNAS, HDAC8, JAG1, KANSL1, KAT6A, KAT6B, KMT2D, MED13, MED13L, NFIX, NIPBL, NR2F1, NR2F2, NSD1, PAX6, RAD21, SATB2, SETD5, SMARCA4, SMARCB1, SMC1A, SOX10, SOX11, SOX2, TBX5, TCF4, |

|  |  |  |  | (deletion), POGZ, TBX5 | TWIST1, WDR26, ZBTB18, ZEB2, ZNF462 |
|---|---|---|---|---|---|
| Neurodevelopmental / Neurological | Autism spectrum disorder, Global developmental delay, Fragile X, Rett, Early infantile epileptic encephalopathy, Dystonia–parkinsonism, Lissencephaly, Ataxia– | Chd8 related neurodevelopmental disorder; Global developmental delay, intellectual disability, ambiguous genitalia, spastic quadriplegia, seizures, abnormal brain mri; Interstitial lung disease, intellectual disability; Nars related disorder; Ppp2r5d neurodevelopment delay developmental disorder; Pten hamartoma syndrome; Shank3 intellectual disability syndrome; Simpson-golabi-behmel syndrome; Slc35a2 congenital disorder of glycosylation; Tetralogy of fallot, autism, pancytopenia | COX20, RNF216, CDK13, MECP2, ATCAY, SPAST, WWOX, FXN, | MYO7A, UNC13D, FMR1, DMD, KCNQ4, MYO15A, TRIOBP, | ADNP, AHDC1, AUTS2, CASK, CC2D1A, CDKL5, CIC, CNOT3, CTCF, CUL3, DDX3X, EBF3, FBXO11, FMR1, FOXG1, GATAD2B, GRIA2, GRIN2A, GRIN2B, HNRNPK, IQSEC2, MECP2, MED12, NAA15, NARS, NUP188, PHF21A, PHIP, POGZ, POU3F3, PTEN, PUS7, RBM10, RERE, SCN1A, SCN2A, SETD1A, SHANK3, SIAH1, SON, SPATA5, SRRM2, SYNGAP1, TAOK1, TCF12, TLK2, TRIO, TRIP12, USP7, WAC, ZDHHC9, ZNF711 |
| Neuromuscular / Skeletal | Duchenne MD, Becker MD, Congenital myopathy, Myasthenic syndrome, Nemaline myopathy, Congenital MD, Centronuclear MD, Osteogenesis imperfecta, Stickler, Spondyloepiphyseal dysplasia, Frontometaphyseal dysplasia | Benign (harmless) duplication of chromosome xp22.31; Cloves syndrome; Gut dysmotity, joint laxity, mast cell disease; Iga deficiency, celiac disease, growth hormone deficiency; Klinefelter syndrome; Marfan syndrome; Neuromuscular disease, progressive kyphoscoliosis, joint contractures, equinovarus deformity; Noonan syndrome | – | MYI7, COL11A1, KMT1, TTN, TGDS, COL4A1, ACTA1, COL2A1, CLCN1, LAMB3, RAF1, TRPV4, DPYD, EIF2AK3, GDFS, ACTC1, MYBPC3 | ACAN, ACTC1, COL11A1, COL1A1, COL1A2, COL2A1, COL3A1, DMD, FBN1, FBN2, FLNA, MYBPC3, MYH7, NF1, PMP22, PORCN, TNNT2, TTN |
| Metabolic / Mitochondrial | – | Cystic fibrosis | DNAH11, KCNJ11, ACSF3, G6PD, KCNJ2, SLC18A2, KCNJ11, COLGALT1, VDDR2, AIFM1, ALP3, ANKRD1, COL4A1, GBE1, SSMID1, | TAI, CYP21A2, PYGL, SRD5, ATP2C1, PLECR1, HPSE1, MTCK, DCR2, RPL5, HPLS1, PQPB1, SLC6A7, SLC26A4 | ABHD5, AGL, ALPL, BCS1L, BTD, CEL, CFTR, COG4, DGUOK, DHCR7, G6PC, G6PD, GBA, GCK, HK1, HNF1A, HNF1B, IBA57, MT-TK, MT-TLL, MTDNA, MTHFR, MVK, ND4, NPHP1, NPHS1, NPHS2, PHKA1, PIGL, PKD1, PKHD1, SLC12A2, |
| | | | LIFR2, MRP53B, DNR, PMM2A, DUOX2, VDR | | SLC26A4, SLC2A1, SLC35A2, TAZ, TGM5, TMEM126A, UROS |
| Hematologic / Immunologic | Diamond–Blackfan anemia, Fanconi anemia, Hemophagocytic lymphohistiocytosis, Shwachman–Diamond anemia, Complement 3 deficiency, Familial Mediterranean fever, Blau syndrome, Nephrotic syndrome, Sickle cell anemia | Castleman disease; Hearing loss, likely 2ndary to chronic active ebv; Iga deficiency; Neutropenia, reticulocytopenia; Sickle cell anemia | – | C8NC, LRBA, SBDS, MEFV, STAT1, C2, SYKL1 | BRCA1, BRCA2, CD40LG, DOCK8, FAS, G6PD, HBB, IKZF1, IL11RA, MEFV, MSH6, NOD2, RET, SBDS, SH2D1A, SLC4A1, SOCS1, STAT1, STAT3, STK11, TNFRSF13B, TNFRSF1A, WAS |
| Connective Tissue / Vascular | Marfan, Cutis laxa, Diffuse pulmonary malformation, Klippel–Trenaunay syndrome, Familial cerebral cavernous malformation, Ehlers–Danlos, Familial hypercholesterolemia | – | – | MED1, RAF1, COL3A1, COL2A1 | APC, AXIN2, COL1A1, COL1A2, COL3A1, CTNNB1, EPHB4, FBN1, FBN2, KRIT1, LDLR, PIK3CA, RASA1, SMAD4, TGFB1 |
| Dermatologic / Ectodermal | Epidermolysis bullosa simplex, Junctional epidermolysis bullosa, Dystrophic epidermolysis bullosa, Lamellar ichthyosis, X-linked ichthyosis, Ectodermal dysplasia, Palmoplantar keratoderma | Evan syndrome; Myhre syndrome; Pendred syndrome; Wolfram | – | GJB2, KRT5, EDA, STS | EBP, EDN3, EDNRB, FLG, GJB2, GJB6, GPR143, KIT, MITF, OCA1, OCA2, PAX3, SOX10, TGM5, WNT10A |
| Cranial / Sensory | Bilateral sensorineural hearing loss, Usher (types 1–2), Otosclerosis, Pendred, Retinitis pigmentosa, Anophthalmia, Persistent Müller duct, Axenfeld–Rieger | – | GP4.3, BC15A3, IFL1, MK3, SCNN1B | GJB2, STRC, OTOF, MYO7A, CDH23, MYO15A, TMC1, ADGRV1, RAC1 | ADGRV1, CDH23, CHD7, EYA1, FOXC1, GJB2, GJB6, KCNQ4, MYO15A, MYO7A, NDP, OPA1, OTOF, PAX6, PCDH15, PITX2, SIX1, SLC26A4, SOX2, STRC, TMPRSS3, USH2A, WFS1 |
| Cardiac / Multisystem | Long QT syndrome, Brugada syndrome, Hypertrophic cardiomyopathy, Oram, Dilated cardiomyopathy, Short QT syndrome, Atrial ARDS, Noonan syndrome, | Danon disease | TNFSF5, FBN1, SCN5A, MYBPC3, TP53, HGF, TNXB, RAC1, BRAF, ARE, | TNNI3, ACTC1, TTN, RAF1, PTPN11 | ACTA2, ACTC1, BRAF, CALM1, CASQ2, CBL, DSC2, DSG2, DSP, GATA4, GATA6, GLA, HRAS, JAG1, JUP, KCNE1, |
| | Alagille syndrome, Prader–Willi, Down syndrome | | ASPA, CAP05 | | KCNE2, KCNH2, KCNQ1, KRAS, LAMP2, LMNA, MAP2K1, MAP2K2, MYBPC3, MYH11, MYH6, MYH7, MYL2, MYL3, NF1, NKX2-5, NOTCH1, NOTCH2, PKP2, PLN, PTPN11, RAF1, RIT1, RYR2, SCN5A, SHOC2, SMAD3, SOS1, SPRED1, TAZ, TBX1, TBX5, TGFBR1, TGFBR2, TMEM43, TNNI3, TNNT2, TPM1, TRDN |
| Other / Syndromic | CHIME, Bardet–Biedl, Marshall–Smith, Burin–Stirling–Limber | Sema3a related syndrome; Smith lemli opitz; Tgm5 related skin fragility disorder etc. | – | – | ABCA3, ABCC9, ABCD1, ADAR, ANO6, ANOS1 etc. |

**Table S5: Methodological Comparison of RAG Approaches.** Different RAG methods are compared based on their design characteristics. Our approaches are designed for text-based rare disease diagnosis, complementing recent multimodal and dynamic RAG methods.

| **Method** | **Input Type** | **Retrieval Strategy** | **Target** | **Key Feature** | **Limitation** |
|---|---|---|---|---|---|
| Standard RAG | Text | Single-shot | General QA | Basic retrieval | Limited to general domains |
| Self-RAG | Text | Adaptive | General | Self-critique | Requires training |
| Dynamic RAG | Text | Dynamic reranking | General QA | RL-based | Complex infrastructure |
| MMed-RAG | Text + Images | Domain-aware | Medical VQA | Multimodal | Requires image data |
| RULE | Text + Images | Calibrated | Medical vision | Risk control | Requires multimodal input |
| RAG-driven CoT (Ours) | Text | Single-shot | Rare disease | Retrieval -> CoT | Text-only |
| CoT-driven RAG (Ours) | Text | Query-refined | Rare disease | CoT -> Retrieval | Text-only |

# Supplementary Notes

**Data Sources and Acquisition for the RAG Knowledge Base**

*HPO Database*

The HPO[1] database can be downloaded directly from the Human Phenotype Ontology website (https://hpo.jax.org/) and then follow the process below to translate the raw HPO ids to textual HPO terms. We convert each HPO id into corresponding term using the HPO mapping dictionary, and then concatenate the resulting terms with semicolons. For example, "HPO: 0000486; HPO: 0001263; HP:0010864" becomes "Strabismus Global developmental delay Intellectual disability severe". The current HPO database contains over 18,000 terms and over 156,000 annotations to hereditary diseases.

*OMIM Database*

To access data from the OMIM[2] (Online Mendelian Inheritance in Man) website (https://www.omim.org/), researchers must first create an account at OMIM and obtain an API key for authentication. Once the key is secured, GET request can be sent, using either Python or Linux terminal, to the OMIM API endpoint at https://api.omim.org/api/entry. When retrieving data for a specific entry, the request should include parameters such as mimNumber, fieldSection, and the API key. For example, a typical request might be structured as: https://api.omim.org/api/entry?mimNumber={omim_id}&include={fieldSection}&apiKey={omim_key}. The API returns structured XML data that details genetic disorders, genes, and phenotypes.

In this study, we focused on extracting key fields that provide valuable insights for large language models interpreting medical information. These fields include clinicalFeatures, description, diagnosis, geneFunction, geneticVariability, genotype, genotypePhenotypeCorrelations, heterogeneity, inheritance, otherFeatures, and phenotype. This curated dataset comprises 10,111 phenotype descriptions (including rare disease diagnoses) and 15,383 gene descriptions, offering a comprehensive resource for Retrieval Augmented Generation (RAG) applications in medical research.

**Case Studies on Impacts of CoT and RAG-based Techniques**

## Case 1: RAG-based Techniques Enhance Top-1 Accuracy on ACTA1 Gene Prioritization

***Clinical Note:***

'The patient, a female, presents with a distinctive clinical profile primarily characterized by hypotonia and areflexia, suggestive of neuromuscular involvement. The hypotonia is manifesting as a reduced muscle tone, observable throughout her musculature, paired with a lack of reflexive responses during neurological examination. She also exhibits proximal muscle weakness, notable in her larger muscle groups which suggests that her muscular impairment is more significant towards the axis of the body. This is coupled with weakness of her facial musculature, evident in a bland facial expression and potential functional challenges with facial activities.\n\nHer oral examination reveals a high palate, a structural anomaly that can be associated with skeletal muscle conditions, potentially affecting feeding and speech. Despite these findings, intellectual development remains within normal limits, excluding intellectual disability as a contributing factor to her presentation. There is an absence of distal muscle weakness, with muscle strength in her hands and feet being relatively preserved, and no evidence of ophthalmoplegia that would indicate ocular muscle involvement. \n\nMotor development has been delayed, as noted in her early childhood records, an element that aligns with her general muscle weakness. Musculoskeletal examination also reveals an ankle flexion contracture, suggesting chronic muscle shortening possibly due to sustained weakness and imbalance. Importantly, there is no history of drooling or bulbar palsy, indicating that the motor issues have not significantly impacted her bulbar function.\n\nThe overall symptomatology matches a clinical entity characterized by muscle weakness starting in childhood, which has been genetically corroborated. The patient's genetic testing revealed a heterozygous variant, consistent with a diagnosis linked to congenital muscle disorder, established through causative genomic interpretation. This condition generally follows an autosomal dominant pattern and is known to present with the described clinical features. The exclusion of significant intellectual impairment and the absence of distal muscle and eye muscle involvement further refine the diagnostic picture in this case.

***Ground Truth:***

ACTA1

***Results of Base Llama3.3-70B-Instruct:***

1. 'MTM1' 2. 'BIN1' 3. 'TNNT1' 4. 'TPM2' 5. 'NEB' 6. **'ACTA1'** 7. 'KBTBD13' 8. 'COL6A1' 9. 'COL6A2' 10. 'COL6A3'"

***Results of Base Llama3.3-70B-Instruct with RAG:***

**1. 'ACTA1'** 2. 'NEB' 3. 'TPM2; 4. 'TPM3' 5. 'TNNT1' 6. 'KBTBD13' 7. 'KLHL40' 8. 'KLHL41' 9. 'LMOD3' 10. 'MYH7'

***Results of Llama3.3-70B-Instruct with CoT-driven RAG:***

Based on the provided clinical note, the following phenotypic features (HPO terms) can be extracted and classified according to the Human Phenotype Ontology (HPO) system:

1. **Hypotonia** - Musculoskeletal system

2. **Areflexia** - Nervous system

3. **Proximal muscle weakness** - Musculoskeletal system

4. **Facial muscle weakness** - Musculoskeletal system

5. **High palate** - Head and neck

6. **Delayed motor development** - Nervous system

7. **Ankle flexion contracture** - Musculoskeletal system

8. **Bland facial expression** - Musculoskeletal system

9. **Normal intellectual development** - This is not an HPO term as it describes the absence of a condition (intellectual disability), but it's crucial for differential diagnosis.

10. **Absence of distal muscle weakness** - This is not an HPO term as it describes the absence of a condition, but it helps in narrowing down the diagnosis.

11. **Absence of ophthalmoplegia** - This indicates no eye muscle involvement, which is important for diagnosis but is described by the absence of a condition.

Given the instruction to only include explicitly stated phenotypic features and to classify each feature under one of the provided categories, the above list represents the direct classifications based on the HPO system. However, for the purpose of this task, we focus on the explicitly mentioned phenotypic abnormalities:

- **Hypotonia**: Musculoskeletal system\n- **Areflexia**: Nervous system\n- **Proximal muscle weakness**: Musculoskeletal system\n- **Facial muscle weakness**: Musculoskeletal system\n- **High palate**: Head and neck\n- **Delayed motor development**: Nervous system\n- **Ankle flexion contracture**: Musculoskeletal system\n- **Bland facial expression**: Musculoskeletal system

And considering the necessity to prioritize genes based on the provided phenotypic features and the clinical context of a congenital muscle disorder with autosomal dominant inheritance, the following genes are relevant for consideration, keeping in mind the final output requires exactly 10 prioritized genes:

1. **ACTA1** - Associated with congenital muscle disorders, including nemaline myopathy, which can present with hypotonia, muscle weakness, and delayed motor development.

2. **RYR1** - Linked to central core disease and other myopathies, presenting with muscle weakness and hypotonia.

3. **SEPN1** - Associated with congenital muscular dystrophy and myopathies, which can include features like hypotonia and delayed motor development.

4. **NEB** - Nemaline myopathy gene, which aligns with the clinical presentation of muscle weakness and hypotonia.

5. **TPM2** - Associated with congenital myopathies, including nemaline myopathy, presenting with hypotonia and muscle weakness.

6. **TNNT1** - Linked to nemaline myopathy, which presents with hypotonia, muscle weakness, and delayed motor development.

7. **MTM1** - Myotubularin 1, associated with X-linked myotubular myopathy, presenting with severe hypotonia and muscle weakness, though the inheritance pattern here is different.

8. **BIN1** - Associated with centronuclear myopathy, which can present with muscle weakness and hypotonia.

9. **DNM2** - Dynamin 2, associated with centronuclear myopathy, presenting with muscle weakness and hypotonia.

10. **KBTBD13** - Associated with nemaline myopathy, which aligns with the clinical features of hypotonia, muscle weakness, and delayed motor development.

These genes are prioritized based on their association with congenital muscle disorders that match the described phenotypic features and the autosomal dominant inheritance pattern mentioned in the clinical note.,

The patient's **age, gender, and ethnicity** can provide valuable context for assessing the likelihood of gene involvement in the observed phenotypes. Here\'s how these factors might influence the consideration of gene involvement:

1. **Age**: The patient\'s age is not explicitly mentioned, but the clinical note refers to "early childhood records" and "childhood," suggesting that the onset of symptoms occurred early in life. This is consistent with congenital muscle disorders, which often present at birth or during early childhood. The age of onset can help narrow down the list of potential genes, as some congenital myopathies have distinct age-related features. For example, nemaline myopathy and centronuclear myopathy often present in infancy or early childhood.

2. **Gender**: The patient is female. While many congenital myopathies do not have a gender bias, some may exhibit a slightly different prevalence or severity between genders due to X-linked inheritance patterns. For instance, myotubular myopathy (associated with the MTM1 gene) is X-linked recessive, predominantly affecting males, although females can be carriers or exhibit milder symptoms due to X-chromosome inactivation patterns. The fact that the patient is female might slightly lower the likelihood of X-linked conditions but does not rule them out entirely, especially if considering autosomal dominant or recessive inheritance patterns.

3. **Ethnicity**: The patient\'s ethnicity is not specified. Ethnicity can influence the likelihood of certain genetic conditions due to founder effects or varying frequencies of specific mutations within different populations. For example, some genetic disorders are more common in specific ethnic groups due to historical genetic bottlenecks or founder effects. Without this information, it\'s challenging to assess how ethnicity might influence the likelihood of specific gene involvement. However, congenital myopathies can occur across all ethnic groups, and their diagnosis is primarily based on clinical features, family history, and genetic testing rather than ethnic background.

Considering these factors, the initial list of genes provided remains relevant, as the clinical presentation of congenital muscle disorder with autosomal dominant inheritance is less likely to be significantly influenced by gender and more by the specific clinical features and age of onset. The genes listed (ACTA1, RYR1, SEPN1, NEB, TPM2, TNNT1, BIN1, DNM2, KBTBD13) are associated with conditions that can present in both genders and across different ethnicities, although specific mutations might have varying frequencies in different populations.\n\nIn summary, while age supports the consideration of congenital myopathies, gender slightly influences the likelihood of X-linked conditions, and ethnicity could potentially impact the frequency of specific mutations but is not provided. The clinical presentation and autosomal dominant inheritance pattern remain the primary guides for gene prioritization in this case.'},

Based on the extracted HPO terms (**Hypotonia**, **Areflexia**, **Proximal muscle weakness**, **Facial muscle weakness**, **High palate**, **Delayed motor development**, **Ankle flexion contracture**, and **Bland facial expression**), the most relevant genes associated with these phenotypes are involved in congenital myopathies and muscular disorders. Here's a refined list of genes, considering the clinical context and the HPO terms:

1. **ACTA1** - Associated with nemaline myopathy, which presents with hypotonia, muscle weakness, and delayed motor development.

2. **RYR1** - Linked to central core disease and other myopathies, presenting with muscle weakness, hypotonia, and potential for delayed motor development.

3. **NEB** - Nemaline myopathy gene, which aligns with the clinical presentation of muscle weakness, hypotonia, and delayed motor development.

4. **TPM2** - Associated with congenital myopathies, including nemaline myopathy, presenting with hypotonia, muscle weakness, and delayed motor development.

5. **TNNT1** - Linked to nemaline myopathy, which presents with hypotonia, muscle weakness, and delayed motor development.

6. **SEPN1** - Associated with congenital muscular dystrophy and myopathies, which can include features like hypotonia and delayed motor development.

7. **BIN1** - Associated with centronuclear myopathy, which can present with muscle weakness, hypotonia, and potential for delayed motor development.

8. **DNM2** - Dynamin 2, associated with centronuclear myopathy, presenting with muscle weakness, hypotonia, and potential for delayed motor development.

9. **KBTBD13** - Associated with nemaline myopathy, which aligns with the clinical features of hypotonia, muscle weakness, and delayed motor development.

10. **MTMR14** - While not initially listed, this gene is associated with myotubular myopathy and centronuclear myopathy, conditions that can present with hypotonia, muscle weakness, and delayed motor development, making it a relevant consideration based on the HPO terms extracted.

These genes are prioritized based on their association with congenital muscle disorders that match the described HPO terms, focusing on conditions that typically present with hypotonia, muscle weakness, and delayed motor development. The specific combination and severity of the HPO terms, along with the clinical context provided, guide the selection of these genes as the most relevant for further investigation.

Based on the extracted HPO terms and the clinical context of a congenital muscle disorder with an autosomal dominant inheritance pattern, the list of genes can be refined by considering inheritance patterns, known pathogenic variants, and functional impact. Here's a refined list:

1. **ACTA1** - Autosomal dominant inheritance, known pathogenic variants cause nemaline myopathy, which presents with hypotonia, muscle weakness, and delayed motor development. Functional impact: Acts as a major component of skeletal muscle thin filaments, crucial for muscle contraction.

2. **RYR1** - Autosomal dominant inheritance, known pathogenic variants cause central core disease and other myopathies, presenting with muscle weakness, hypotonia, and potential for delayed motor development. Functional impact: Plays a critical role in releasing calcium ions from the sarcoplasmic reticulum, essential for muscle contraction.

3. **TPM2** - Autosomal dominant inheritance, known pathogenic variants cause congenital myopathies, including nemaline myopathy, presenting with hypotonia, muscle weakness, and delayed motor development. Functional impact: Encodes a tropomyosin isoform important for the regulation of muscle contraction.

4. **TNNT1** - Autosomal recessive inheritance, known pathogenic variants cause nemaline myopathy, which presents with hypotonia, muscle weakness, and delayed motor development. However, considering the autosomal dominant pattern in the clinical note, this might be less likely unless there's a de novo mutation or an atypical inheritance pattern. Functional impact: Essential for the regulation of muscle contraction as part of the troponin complex.

5. **KBTBD13** - Autosomal dominant inheritance, known pathogenic variants cause nemaline myopathy, which aligns with the clinical features of hypotonia, muscle weakness, and delayed motor development. Functional impact: Acts as a component of the BTB/Kelch family, potentially involved in the regulation of muscle growth and development.

6. **BIN1** - Autosomal dominant inheritance, known pathogenic variants cause centronuclear myopathy, which can present with muscle weakness, hypotonia, and potential for delayed motor development. Functional impact: Involved in the regulation of muscle cell membrane dynamics and potentially in the pathogenesis of centronuclear myopathy.

7. **DNM2** - Autosomal dominant inheritance, known pathogenic variants cause centronuclear myopathy, presenting with muscle weakness, hypotonia, and potential for delayed motor development. Functional impact: Encodes dynamin 2, crucial for membrane trafficking and potentially involved in the pathogenesis of centronuclear myopathy.

8. **NEB** - Autosomal recessive inheritance, known pathogenic variants cause nemaline myopathy, which presents with hypotonia, muscle weakness, and delayed motor development. Like TNNT1, this might be less likely given the autosomal dominant pattern unless considering a de novo mutation or atypical inheritance. Functional impact: Essential for the structure and function of skeletal muscle thin filaments.

9. **SEPN1** - Autosomal recessive inheritance, known pathogenic variants cause congenital muscular dystrophy and myopathies, which can include features like hypotonia and delayed motor development. Given the autosomal dominant pattern in the clinical note, this is less likely unless there's an atypical inheritance pattern. Functional impact: Involved in the regulation of calcium homeostasis and potentially in the pathogenesis of congenital muscular dystrophy.

10. **MYPN** - Autosomal dominant inheritance, known pathogenic variants cause nemaline myopathy and other myopathies, presenting with hypotonia, muscle weakness, and delayed motor development. Functional impact: Encodes a protein involved in the regulation of muscle Z-disk structure and function.\n\nThe refined list prioritizes genes with autosomal dominant inheritance patterns and known pathogenic variants that cause congenital muscle disorders matching the patient's HPO terms. The functional impact of each gene on muscle structure and function is also considered, highlighting their potential role in the pathogenesis of the observed phenotypes."

**1. 'ACTA1'** 2. 'RYR1' 3. 'TPM2' 4. 'KBTBD13' 5. 'BIN1' 6. 'DNM2' 7. 'MYPN' 8. 'NEB' 9. 'TNNT1'

10. 'CCDC174'

***Results of Llama3.3-70B-Instruct with RAG-driven CoT:***

Step 1: Extract and classify HPO terms

The patient's clinical profile includes hypotonia, areflexia, proximal muscle weakness, high palate, delayed motor development, ankle flexion contracture, and absence of distal muscle weakness or ophthalmoplegia. Classifying these under the given categories:

- Musculoskeletal system: hypotonia, areflexia, proximal muscle weakness, ankle flexion contracture

- Head and neck: high palate

- Nervous system: delayed motor development

- Constitutional symptoms: none

- Growth abnormality: none

Other categories are not directly relevant based on the provided information.

Step 2: Assess demographic impact

The patient's age, gender, and ethnicity are not explicitly mentioned, but the clinical presentation suggests a congenital or early-onset condition. The absence of specific demographic details limits the assessment of their impact on gene involvement. However, the condition described is consistent with an autosomal dominant pattern, which does not have a gender bias.

Step 3: Map to relevant gene-disease associations

Based on the extracted HPO terms, genes associated with congenital muscle disorders, particularly those involving hypotonia, areflexia, and proximal muscle weakness, are relevant. Conditions such as nemaline myopathy, congenital fiber-type disproportion, and other myopathies could be considered.

Step 4: Refine based on inheritance patterns and variant evidence \nGiven the autosomal dominant inheritance pattern and the presence of a heterozygous variant, genes known to cause dominant congenital myopathies are prioritized. The clinical features and genetic evidence point towards genes involved in the structure and function of muscle fibers.

Step 5: Prioritize the top 10 genes

Considering the clinical presentation and genetic information, genes associated with congenital myopathies and muscle disorders are prioritized. The specific genes involved in the structure and function of muscle fibers, and known to follow an autosomal dominant inheritance pattern, are selected.

**1. 'ACTA1'** 2. 'NEB' 3. 'TPM3' 4. 'TNNT1' 5. 'KBTBD13' 6. 'KLHL40' 7. 'LMOD3' 8. 'MYH7'

9. 'CCDC78' 10. 'MYPN'

## ***Case 2: CoT Prompting Enhances Top-1 Accuracy on ZIC3 Gene Prioritization***

***Clinical Note:***

The patient, a male, presents with several significant congenital cardiac anomalies and splenic abnormalities. From birth, he has exhibited a single ventricle heart defect, a complex condition where one of the ventricles is underdeveloped or absent, affecting normal circulatory function. Alongside this, he is affected by polysplenia, a condition characterized by the presence of multiple small accessory spleens instead of a single, normal-sized spleen. Fortunately, asplenia, the complete absence of splenic tissue, has been ruled out in this patient.\n\nFurther complicating his clinical picture is the congenital presence of dextrotransposition of the great arteries (d-TGA), where the positions of the main arteries, namely the aorta and the pulmonary artery, are switched, resulting in significant circulatory issues if not corrected. Additionally, the patient has pulmonary valve atresia, a defect where the pulmonary valve does not form properly, impeding blood flow from the right ventricle to the lungs, which further impacts oxygenation of the blood.\n\nDespite these multiple cardiac anomalies, the patient does not exhibit mitral atresia, suggesting that the mitral valve, which regulates blood flow between the left atrium and left ventricle, is present and functional to some extent.\n\nThis comprehensive clinical presentation is consistent with a diagnosis that has been confirmed as an underlying genetic condition, which typically emerges from birth. The diagnosis has been further substantiated at the genetic level, confirming the presence of a causative mutation responsible for the array of presented phenotypes. This condition reflects a complex interplay of anomalies that indicate a significant impact on the patient's cardiac function and splenic morphology from an early stage of life. The clinical course necessitates a multi-disciplinary approach to address these congenital issues, with potential surgical and medical interventions to manage symptoms and improve quality of life.

***Ground Truth:***

ZIC3

***Results of Base Llama3.3-70B-Instruct:***

POTENTIAL_GENES:\n1. 'NKX2-5'\n2. 'GATA4'\n3. 'TBX5'**\n4. 'ZIC3'**\n5. 'CRELD1'\n6. 'GJA1'\n7. 'TFAP2B'\n8. 'FOXA2'\n9. 'PKD2'\n10. 'NOTCH1'

***Results of Base Llama3.3-70B-Instruct with CoT Prompting:***

<|start_header_id|>assistant<|end_header_id|>\n\nREASONING: \nStep 1: Extract and classify HPO terms - The patient\'s clinical presentation includes single ventricle heart defect, polysplenia, dextrotransposition of the great arteries (d-TGA), and pulmonary valve atresia. These terms can be classified under the "Cardiovascular system" and "Immune system" categories due to the involvement of the heart and spleen. \nStep 2: Assess demographic impact - The patient is a male, and while gender can influence the likelihood of certain genetic conditions, the provided information does not specify how gender impacts the current condition directly. Age is not explicitly mentioned but given the congenital nature of the conditions, it\'s clear the patient has been affected from birth. Ethnicity is not provided. \nStep 3: Map to relevant gene-disease associations - Genes associated with cardiac development and splenic formation are of interest. Conditions like heterotaxy syndrome, which involves abnormalities in the formation of asymmetric organs such as the heart and spleen, are relevant. \nStep 4: Refine based on inheritance patterns and variant evidence - The inheritance pattern is not explicitly stated, but given the complexity and congenital nature of the defects, it\'s likely an autosomal dominant or X-linked condition. Genes known to cause similar phenotypes when mutated, such as those involved in the NOTCH and hedgehog signaling pathways, should be considered. \nStep 5: Prioritize the top 10 genes - Based on the involvement of cardiac and splenic anomalies, genes known to be crucial in the development of these systems will be prioritized. This includes genes associated with laterality defects and cardiac morphogenesis.\

POTENTIAL_GENES:

**1. \'ZIC3\'** 2. \'NKX2-5\' 3. \'GATA4\' 4. \'NOTCH1\' 5. \'CRELD1\' 6. \'LEFTY2\' 7. \'ACVR2B\' 8. \'NODAL\' 9. \'FOXH1\' 10. \'TBX1\''

***Results of Base OpenAI GPT-4o:***

POTENTIAL_GENES:

1. 'NKX2-5' 2. 'GATA4' **3. 'ZIC3'** 4. 'TBX5' 5. 'CRELD1' 6. 'CFC1' 7. 'NODAL' 8. 'ACVR2B'

9. 'SHH' 10. 'LEFTY2'

***Results of Base OpenAI GPT-4o with CoT Prompting:***

REASONING:

Step 1: Extract and classify HPO terms

- Cardiovascular system:
  - HP:0001627 (Congenital heart disease)
  - HP:0001638 (Single ventricle)
  - HP:0002566 (Dextro-transposition of the great arteries)
  - HP:0004933 (Pulmonary valve atresia)
- Immune system:
  - HP:0010454 (Polysplenia)
- Prenatal development or birth:
  - HP:0001744 (Congenital anomaly)

Step 2: Assess demographic impact

- The patient is male, which does not strongly influence genetic predisposition to these conditions.
- The congenital nature of the anomalies suggests an early developmental gene involvement, likely related to left-right patterning and heart morphogenesis.
- Ethnicity is not specified, but some syndromes such as heterotaxy are seen across populations.

Step 3: Map to relevant gene-disease associations

- Genes associated with heterotaxy and congenital heart defects:
    - ZIC3 (X-linked heterotaxy)
    - NODAL (left-right patterning)
    - CFC1 (cardiac and laterality defects)

- NKX2-5 (congenital heart defects)
- GATA4 (cardiac morphogenesis)
- LEFTY2 (left-right asymmetry defects)
- SHH (developmental patterning)
- ACVR2B (cardiovascular development)
- CRELD1 (atrioventricular septal defects, heterotaxy)
- TXNDC3 (cardiac looping defects)

Step 4: Refine based on inheritance patterns and variant evidence

- X-linked: ZIC3 (male predominance, heterotaxy syndromes)
- Autosomal dominant with variable expressivity: GATA4, NKX2-5, CFC1
- Autosomal recessive: NODAL, LEFTY2, ACVR2B (important for left-right patterning defects)
- De novo or rare dominant: SHH, CRELD1

Step 5: Prioritize the top 10 genes

POTENTIAL_GENES:

1. 'ZIC3'
2. 'NODAL'
3. 'CFC1'
4. 'NKX2-5'
5. 'GATA4'
6. 'LEFTY2'
7. 'SHH'
8. 'ACVR2B'
9. 'CRELD1'
10. 'TXNDC3'